\documentclass[10pt,twocolumn,letterpaper]{article}

\usepackage[pagenumbers]{cvpr} 
\usepackage{cuted}   
\usepackage{capt-of} 
\usepackage{float}
\usepackage{graphicx}
\usepackage{caption} 
\usepackage[table]{xcolor}
\usepackage{stfloats}
\usepackage{balance}
\usepackage{booktabs}      
\usepackage{multirow}      
\usepackage{makecell}      
\usepackage{adjustbox}     
\usepackage{array}         
\usepackage{amsmath}       
\usepackage{amssymb}       
\usepackage{xcolor}        
\usepackage{pifont}
\usepackage{multicol}
\usepackage{caption} 

\newcommand{\cmark}{\ding{51}} 
\newcommand{\xmark}{\ding{55}} 

\usepackage{amssymb}

\definecolor{cvprblue}{rgb}{0.21,0.49,0.74}
\usepackage[pagebackref,breaklinks,colorlinks,allcolors=cvprblue]{hyperref}
\title{
    Toward Generalizable Deblurring: Leveraging Massive Blur Priors with Linear Attention for Real-World Scenarios
}

\author{
    \begin{minipage}[t]{0.31\textwidth}
        \centering
        Yuanting Gao\thanks{Equal contribution. \quad $\dagger$ Corresponding authors. \quad$\triangle$Project lead. \\ \quad \textsuperscript{$\S$}This work was done during his internship at Shanghai AI Laboratory.} \textsuperscript{$\,\S$}\\
        Tsinghua University\\
        {\tt\small gaoyt24@mails.tsinghua.edu.cn}
    \end{minipage}
    \hfill
    \begin{minipage}[t]{0.31\textwidth}
        \centering
        Shuo Cao\textsuperscript{\small\textasteriskcentered} \\
        USTC, Shanghai AI Lab\\
        {\tt\small caoshuo@pjlab.org.cn}
    \end{minipage}
    \hfill
    \begin{minipage}[t]{0.31\textwidth}
        \centering
        Xiaohui Li \\
        SJTU, Shanghai AI Lab\\
        {\tt\small lixiaohui@pjlab.org.cn}
    \end{minipage}
    \vspace{10pt} \\ 
    \begin{minipage}[t]{0.31\textwidth}
        \centering
        Yuandong Pu \\
        SJTU, Shanghai AI Lab\\
        {\tt\small puyuandong@pjlab.org.cn}
    \end{minipage}
    \hfill
    \begin{minipage}[t]{0.31\textwidth}
        \centering
        Yihao Liu\textsuperscript{$\dagger$}\textsuperscript{$\triangle$} \\
        Shanghai AI Lab\\
        {\tt\small liuyihao@pjlab.org.cn}
    \end{minipage}
    \hfill
    \begin{minipage}[t]{0.31\textwidth}
        \centering
        Kai Zhang\textsuperscript{$\dagger$}\\
        Tsinghua University\\
        {\tt\small zhangkai@sz.tsinghua.edu.cn}
    \vspace{-10pt}
    \end{minipage}
}

\begin{document}

\maketitle

\setlength{\stripsep}{-40pt}

\begin{strip}

    \centering
    \includegraphics[width=0.75\linewidth]{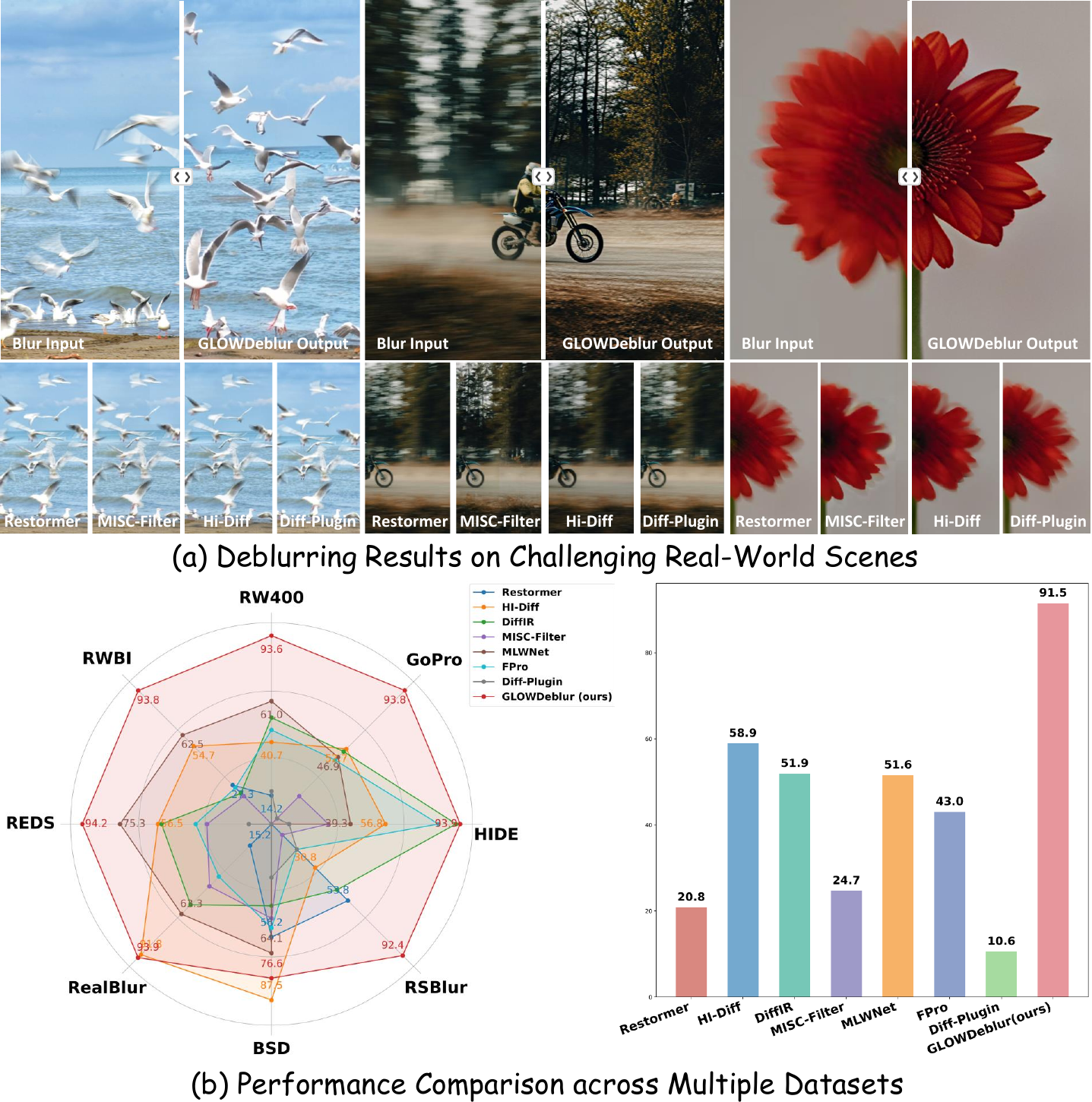}

    \captionof{figure}{(a) \textbf{Visual comparison on challenging real-world images:} our GLOWDeblur effectively restores a wide range of blur patterns, while prior methods often fail in complex scenarios. (b) \textbf{Quantitative comparison on diverse benchmarks:} the left plot shows dataset scores computed by ranking methods on each metric and averaging across metrics; the right plot reports average model scores across all datasets, highlighting the strong generalization ability of GLOWDeblur.}
    \label{fig:teaser}
\end{strip}

\clearpage
\begin{abstract}
Image deblurring has advanced rapidly with deep learning, yet most methods exhibit poor generalization beyond their training datasets, with performance dropping significantly in real-world scenarios. Our analysis shows this limitation stems from two factors: datasets face an inherent trade-off between realism and coverage of diverse blur patterns, and algorithmic designs remain restrictive, as pixel-wise losses drive models toward local detail recovery while overlooking structural and semantic consistency, whereas diffusion-based approaches, though perceptually strong, still fail to generalize when trained on narrow datasets with simplistic strategies. Through systematic investigation, we identify blur pattern diversity as the decisive factor for robust generalization and propose Blur Pattern Pretraining (BPP), which acquires blur priors from simulation datasets and transfers them through joint fine-tuning on real data. We further introduce Motion and Semantic Guidance (MoSeG) to strengthen blur priors under severe degradation, and integrate it into \textbf{GLOWDeblur}, a \textbf{G}eneralizable rea\textbf{L}-w\textbf{O}rld light\textbf{W}eight \textbf{Deblur} model that combines convolution-based pre-reconstruction \& domain alignment module with a lightweight diffusion backbone. Extensive experiments on six widely-used benchmarks and two real-world datasets validate our approach, confirming the importance of blur priors for robust generalization and demonstrating that the lightweight design of GLOWDeblur ensures practicality in real-world applications. The project page is available at \url{https://vegdog007.github.io/GLOWDeblur_Website/}.
\end{abstract}
    
\section{Introduction}
\label{sec:intro}
\begin{figure}[H]
    \centering
    \includegraphics[width=1\linewidth]{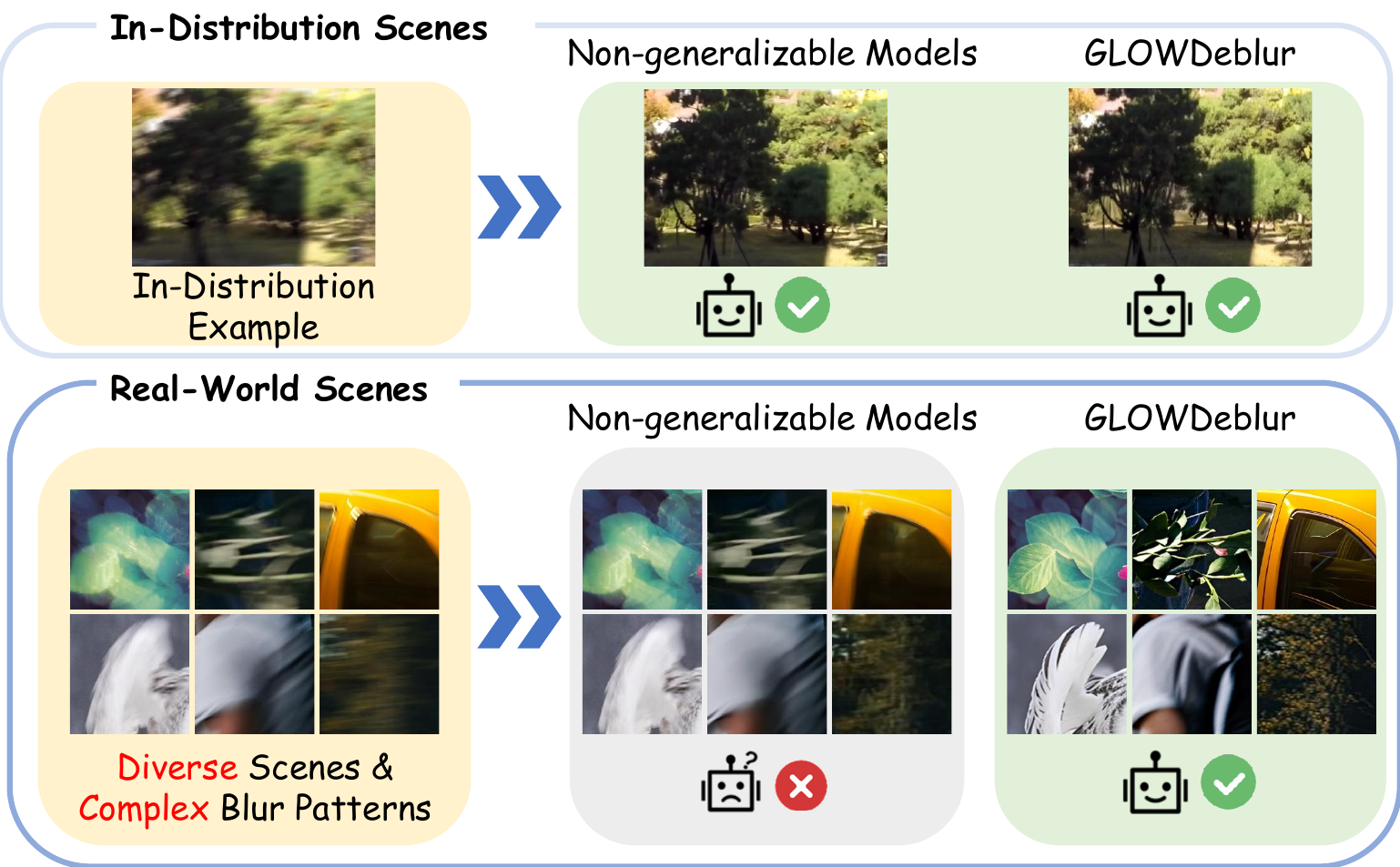}
    \caption{Challenges for Real-World Generalization}
    \label{fig:intro}
\end{figure}

In recent years, image deblurring has made significant progress with the rapid development of deep learning. A variety of high-quality datasets~\cite{gopro,hide,reds,realblur,bsd,rsblur,gsblur} and advanced algorithms~\cite{hi-diff,miscfilter,mlwnet} have been proposed, achieving impressive performance across benchmarks. However, these advances have not resolved a central limitation: most approaches are trained and evaluated on a limited set of datasets, leading to overfitting to their domain characteristics and specific blur patterns. As a result, their generalization performance drops noticeably when applied to real-world scenarios, where blur is inherently more diverse and complex. As illustrated in Figure.~\ref{fig:teaser}, where three real-world cases show that current state-of-the-art methods fail to deliver satisfactory restorations not only in complex scenes but also in a visually simple case, reflecting the inherent challenges of real-world blur. Moreover, substantial gaps exist among current datasets, and naive mixed-dataset training not only fails to improve generalization but often degrades performance on the original benchmarks. This raises a central challenge: how to effectively organize existing datasets and design deblurring frameworks that can substantially improve generalization, enabling models to robustly handle the diverse and complex blur patterns encountered in real-world conditions.


Through systematic investigation, we find that this limitation arises from two key aspects: dataset construction and algorithmic design. Current datasets face inherent constraints, making it difficult to achieve both realism and comprehensive coverage of blur patterns. Synthetic datasets such as GoPro~\cite{gopro} and REDS~\cite{reds} allow large-scale training but diverge from real-world distributions, while real-captured datasets like RealBlur~\cite{realblur} and RSBlur~\cite{rsblur} improve realism but remain limited in blur diversity and scene coverage. Even simulation-based datasets such as GSBlur~\cite{gsblur} still differ significantly from real-world degradations. Consequently, substantial gaps remain both across datasets and between synthetic datasets and real-world blur, hindering models trained on a single dataset from achieving robust generalization. Beyond the data, algorithmic choices also impose important constraints. Models trained with pixel-wise losses (e.g., MSE) favor local details but overlook global structure and semantics, leading to smooth outputs with poor generalization~\cite{restormer,miscfilter,mlwnet,fpro}. Diffusion models leverage strong priors for perceptually better results, but training on narrow datasets with simple strategies limits their ability to capture diverse blur patterns~\cite{hi-diff,diffir,diff-plugin}.

Based on these observations, we first conduct a systematic analysis of dataset biases in deblurring. While prior research has largely emphasized the realism of blur~\cite{realblur,bsd,rsblur}, we find that the diversity and coverage of blur patterns—such as their orientation and spatial distribution—are critical factors behind the gaps observed both across datasets and between datasets and real-world blur. Motivated by this finding, we propose \textbf{BPP} (\textbf{B}lur \textbf{P}attern \textbf{P}retraining): a data-centric strategy where models are first pretrained on large-scale simulation datasets with comprehensive blur patterns to acquire strong blur priors, and are then jointly fine-tuned on real-captured datasets. This process enables the model to leverage blur priors to bridge dataset gaps, ultimately improving both robustness and applicability in real-world deblurring.

In terms of algorithm design, diffusion models offer strong prior modeling and the ability to integrate heterogeneous data sources, making them well suited for generalizable deblurring. However, their high complexity and resource demands hinder deployment in real-world applications that require real-time efficiency, such as autonomous driving and mobile photography. To address this, we propose \textbf{GLOWDeblur}, a \textbf{G}eneralizable rea\textbf{L}-w\textbf{O}rld light\textbf{W}eight \textbf{Deblur} model that combines a convolution-based pre-reconstruction  \& domain-alignment module with a lightweight diffusion model, which employs a Deep Compression AutoEncoder and Linear Attention. To further strengthen the model’s ability to handle diverse and complex real-world blur, we incorporate motion guidance and cross-modal semantic captions as complementary signals, enabling the model to better adapt to varied blur patterns and recover severely degraded regions by leveraging the generative capacity of diffusion models. GLOWDeblur is trained with our Blur Pattern Pretraining (BPP) strategy and extensively evaluated on six widely used benchmarks and two real-world datasets. Since our work targets perceptual quality in diverse real-world blur, we employ a broad set of perceptual-oriented metrics, consistent with recent trends~\cite{supir,linearsr,deqa_score,artimuse} emphasizing visual fidelity beyond pixel-wise accuracy. Results show that GLOWDeblur achieves superior cross-dataset and real-world generalization, underscoring blur priors as the key to real-world deblurring.

\begin{table*}[!b]
\centering
\caption{Cross-dataset results (PSNR/SSIM) reveal severe generalization gaps, with \textcolor{red}{red} indicating the best in-dataset and \textcolor{blue}{blue} the second-best cross-dataset result. Avg. column reports mean PSNR/SSIM across datasets.}
\label{tab:cross_dataset_res}
\resizebox{1\textwidth}{!}{
\begin{tabular}{l|c|c|c|c|c|c|c|c}
\toprule[1.5pt]
\textbf{Training Set} \textbackslash \textbf{Test Set} & \textbf{GoPro}~\cite{gopro} & \textbf{HIDE}~\cite{hide} & \textbf{REDS}~\cite{reds} & \textbf{RealBlur}~\cite{realblur} & \textbf{BSD}~\cite{bsd} & \textbf{RSBlur}~\cite{rsblur} & \textbf{GSBlur}~\cite{gsblur} & \textbf{Avg.} \\
\hline
\textbf{GoPro (Synthetic)}   & \textcolor{red}{32.92 / 0.94} & \textcolor{blue}{31.22 (↓0.40)/ 0.92}  & 26.93 (↓7.46) / 0.82 & 28.96 (↓3.13) / 0.88 & 24.43 (↓9.32) / 0.90 & 29.30 (↓3.68) / 0.86 & 24.94 (↓6.43) / 0.82 & 28.39 / 0.88 \\
\textbf{HIDE (Synthetic)}    & \textcolor{blue}{32.60 (↓0.32) / 0.94}  & \textcolor{red}{31.62 / 0.93} & 26.68 (↓7.71) / 0.83 & 27.76 (↓4.33) / 0.86 & 25.34 (↓8.41) / 0.85 & 27.77 (↓5.21) / 0.83 & 23.40 (↓7.97) / 0.80 & 27.88 / 0.86 \\
\textbf{REDS (Synthetic)}    & 26.21 (↓6.71) / 0.83 & 24.42 (↓7.20) / 0.80 & \textcolor{red}{34.39 / 0.94} & 28.72 (↓3.37) / 0.86 & 28.90 (↓4.85) / 0.84 & 28.09 (↓4.89) / 0.87 & 24.42 (↓6.95) / 0.80 & 27.88 / 0.85 \\
\textbf{RealBlur (Real)}     & 24.50 (↓8.42) / 0.82 & 23.60 (↓8.02) / 0.81 & 25.85 (↓8.54) / 0.79 & \textcolor{red}{32.09 / 0.92} & 28.78 (↓4.97) / 0.91 & 29.65 (↓3.33) / 0.87 & 24.92 (↓6.45) / 0.81 & 27.06 / 0.85 \\
\textbf{BSD (Real)}          & 27.27 (↓5.65) / 0.86 & 26.27 (↓5.35) / 0.85 & 28.20 (↓6.19) / 0.84 & 29.64 (↓2.45) / 0.89 & \textcolor{red}{33.75 / 0.96} & 30.45 (↓2.53) / 0.89 & 26.78 (↓4.59) / 0.85 & 28.91 / 0.88 \\
\textbf{RSBlur (Real)}       & 27.55 (↓5.37) / 0.87 & 25.79 (↓5.83) / 0.84 & 28.08 (↓6.31) /0.84 & \textcolor{blue}{30.41 (↓1.68) / 0.89} & 30.85 (↓2.90) / 0.94 & \textcolor{red}{32.98 / 0.93} & \textcolor{blue}{27.63 (↓3.74) / 0.86} & \textcolor{blue}{29.04} / 0.88 \\
\textbf{GSBlur (Simulated)}  & 28.51 (↓4.41) / 0.90 & 26.12 (↓5.50) / 0.87 & \textcolor{blue}{30.29 (↓4.10) / 0.90} & 30.06 (↓2.03) / 0.91 & \textcolor{blue}{31.24 (↓2.51) / 0.94} & \textcolor{blue}{32.01 (↓0.97) / 0.92} & \textcolor{red}{31.37 / 0.92} & \textcolor{red}{29.94 / 0.91} \\
\bottomrule[1.5pt]
\end{tabular}}
\end{table*}

In this paper, we present the following contributions:

\noindent\textbf{Revealing the role of blur patterns.} We systematically analyze dataset biases and reveal that the diversity and coverage of blur patterns, rather than realism alone, are the decisive factors behind cross-dataset gaps. Learning blur priors and leveraging them as guidance is shown to be essential for achieving robust and quantifiable generalization.

\noindent\textbf{Data and model level priors for generalization.} We introduce Blur Pattern Pretraining (BPP), a data-centric strategy that first learns blur priors from large-scale simulation datasets and then jointly fine-tunes on real-captured datasets. In parallel, we propose Motion and Semantic Guidance (MoSeG) to reinforce blur priors and alleviate structural and semantic degradation under severe blur.

\noindent\textbf{A generalizable real-world deblurring model.} We propose GLOWDeblur, a diffusion-based framework that balances efficiency and effectiveness, achieving strong performance across six benchmarks and two real-world datasets. Beyond results, it also serves as a practical testbed to validate our insights and demonstrate real-world applicability.


\section{Motivation}
\label{sec:formatting}

\subsection{Limitations of Existing Models in Real-World Blur Scenarios}
Although recent methods have achieved remarkable progress, they still exhibit fundamental limitations, particularly in generalizing to diverse real-world blur patterns.
As illustrated in Figure.~\ref{fig:teaser}, across three representative real-world scenes, current state-of-the-art methods fail to deliver satisfactory restorations beyond the training distribution, not only under complex scenes but even in visually simple ones. Figure.~\ref{fig:intro} further reinforces this observation: although existing methods handle in-distribution blur reasonably well, they suffer severe failures when confronted with the diverse and complex scenes and blur patterns of real-world scenarios. This indicates that current approaches rely heavily on dataset-specific distributions rather than learning transferable representations of blur.


These observations motivate us to examine the roots of the generalization gap, revealing that explicitly modeling blur-pattern priors and organizing training data to capture their diversity are crucial for robust real-world deblurring. Guided by these insights, we design improved training strategies and a lightweight model that generalize effectively across diverse scenes and blur patterns (Figure.~\ref{fig:intro} GLOWDeblur), thereby overcoming the limitations of existing methods.
\subsection{Dataset Bias and Blur Pattern Discrepancies}
To understand the generalization gap, we conducted a series of cross-dataset experiments using Restormer as a representative backbone. Models were first trained individually on six widely used datasets and one simulation-based dataset constructed via 3D Gaussian Splatting, and then evaluated across all datasets. As shown in Table.~\ref{tab:cross_dataset_res}, models trained on one dataset degrade notably on others, underscoring a substantial cross-dataset distribution gap.

\begin{figure}[H]
    \centering
    \includegraphics[width=1\linewidth]{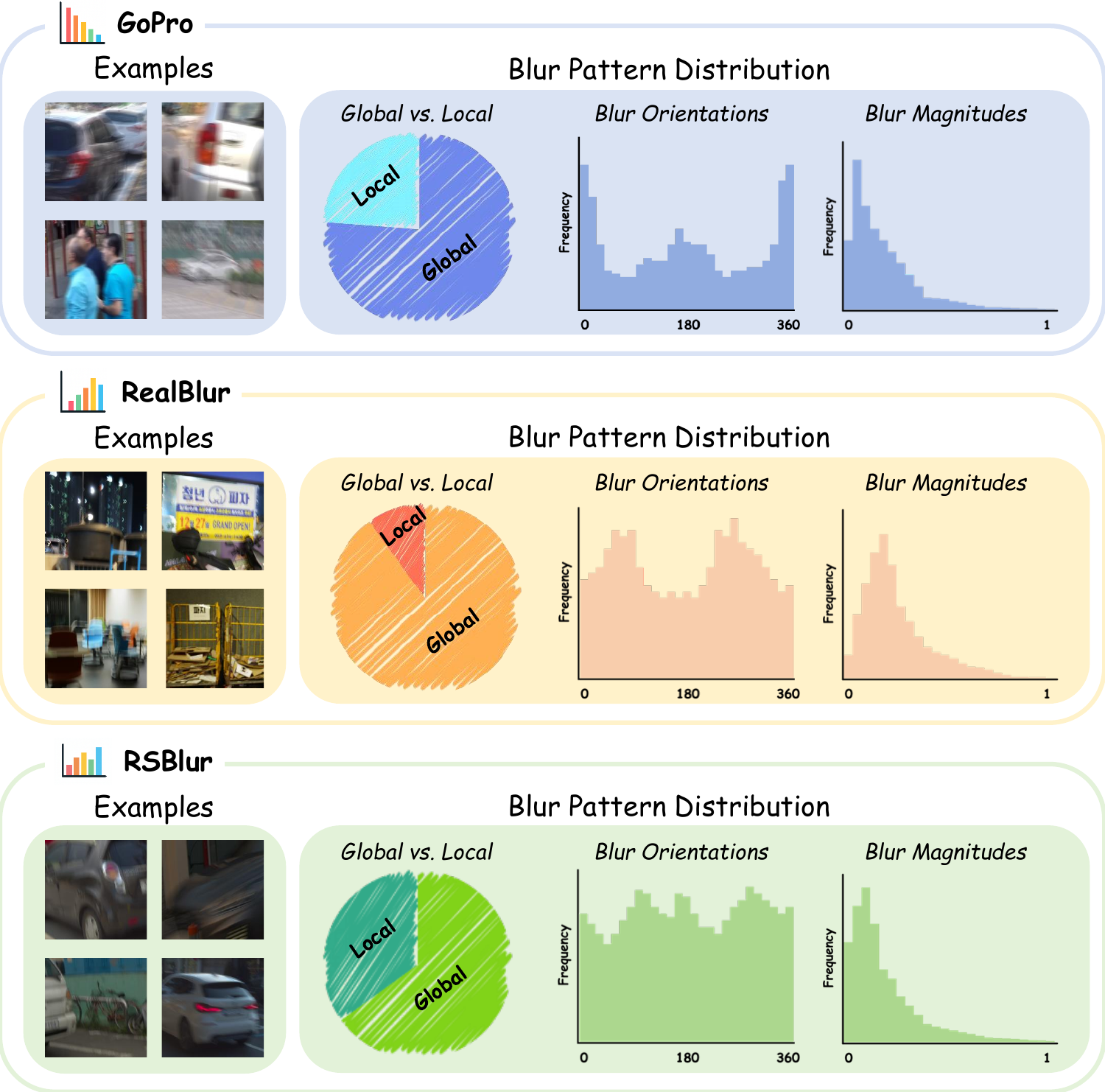}
    \caption{Illustration of dataset-specific blur patterns, highlighting notable distribution differences.}
    \label{fig:blur_pattern_comparison}
\end{figure}

Importantly, Our cross-dataset experiments further reveal two important observations. First, these gaps exist not only between synthetic and real datasets, but also within the same category (synthetic vs. synthetic or real vs. real), indicating that beyond realism there exist deeper sources of mismatch. Second, despite limited realism in both scenes and blur in GSBlur, its broad coverage of blur patterns allows models trained on it to achieve relatively stronger cross-dataset robustness. Collectively, these results highlight that blur pattern diversity, insufficiently recognized in prior work, plays a dominant role in causing the significant cross-dataset gap.

To validate this insight, we conducted a fine-grained analysis of blur characteristics across datasets. As shown in Figure.~\ref{fig:blur_pattern_comparison}, their blur patterns differ markedly in orientation, magnitude, and locality. In particular, GoPro is dominated by horizontal blur while RealBlur is primarily vertical, yet prior work has often attributed their discrepancy only to differences in realism.

In summary, our analysis shows that dataset bias in deblurring arises primarily from blur pattern mismatches, thereby motivating our exploration of both data-centric strategies to mitigate cross-dataset gaps and algorithmic frameworks that exploit blur priors for robust real-world generalization.

\section{Methodology}

\subsection{Blur Pattern Pretraining (BPP)}
Since blur pattern diversity is pivotal for generalization, we propose Blur Pattern Pretraining (BPP), which utilizes datasets with broad blur coverage to enable models to learn intrinsic blur priors, thereby mitigating distribution gaps and enhancing both performance and generalization. Table.~\ref{tab:performance_comparison} empirically validates the effectiveness of this strategy using Restormer as a testbed. Specifically, in settings (a), (b), and (c), applying BPP on a dataset with diverse blur patterns prior to fine-tuning on target real-world datasets (RealBlur, BSD, and RSBlur) consistently yields superior in-distribution accuracy and cross-domain robustness compared to direct training. Furthermore, the comparison reveals the pitfalls of naïve mixed training (d); simply combining datasets fails to achieve optimal results and even degrades performance due to significant domain shifts and conflicting distributions. In contrast, strategy (e) demonstrates that BPP serves as a critical bridge—effectively harmonizing these gaps before mixing, which results in comprehensive improvements across all benchmarks.

\begin{table*}[!t]
\centering
\caption{Performance comparison (PSNR/SSIM) on RealBlur-J, BSD, and RSBlur. Settings (a)-(c) validate BPP's gains in accuracy and robustness. Furthermore, comparing (d) and (e) shows BPP resolves the domain conflicts of naïve mixed training, ensuring comprehensive improvements.}
\label{tab:performance_comparison}
\vspace{-5pt}
\resizebox{1\textwidth}{!}{
\begin{tabular}{c|l|c|c|c|c}
\toprule[1.5pt]
\textbf{No.} & \textbf{Training set} & \textbf{BPP} & \textbf{RealBlur-J}~\cite{realblur} & \textbf{BSD}~\cite{bsd} & \textbf{RSBlur}~\cite{rsblur} \\
\hline
(a) & RealBlur                & \checkmark & \textcolor{red}{32.26 (↑0.17) / 0.93 (↑0.01)} & 29.76 (↓3.99) / 0.92 (↓0.04) & 30.28 (↓2.70) / 0.89 (↓0.04) \\
(b) & BSD                     & \checkmark & 29.95 (↓2.14) / 0.90 (↓0.02) & \textcolor{red}{34.21 (↑0.46) / 0.96 (±0.00)} & 31.15 (↓1.83) / 0.90 (↓0.03) \\
(c) & RSBlur                  & \checkmark & 30.63 (↓1.46) / 0.90 (↓0.02) & 31.22 (↓2.53) / 0.95 (↓0.01) &  \textcolor{red}{33.69 (↑0.71) / 0.94 (↑0.01)} \\
(d) & RealBlur + BSD + RSBlur & $\times$ & 30.83 (↓1.26) / 0.89 (↓0.03) & 31.99 (↓1.76) / 0.95 (↓0.01) & 31.24 (↓1.74) / 0.90 (↓0.03) \\
(e) & RealBlur + BSD + RSBlur & \checkmark & \textcolor{blue}{32.11 (↑0.02) / 0.93 (↑0.01)} & \textcolor{blue}{33.62 (↓0.13) / 0.96 (±0.00)} & \textcolor{blue}{33.65 (↑0.67) / 0.94 (↑0.01)} \\
\hline
\rowcolor{gray!20} & \textbf{Best same-dataset performance} & -- & \textbf{32.09 / 0.92} & \textbf{33.75 / 0.96} & \textbf{32.98 / 0.93} \\
\bottomrule[1.5pt]
\end{tabular}}
\end{table*}

\begin{figure*}[!b]
    \centering
    \includegraphics[width=0.8\linewidth]{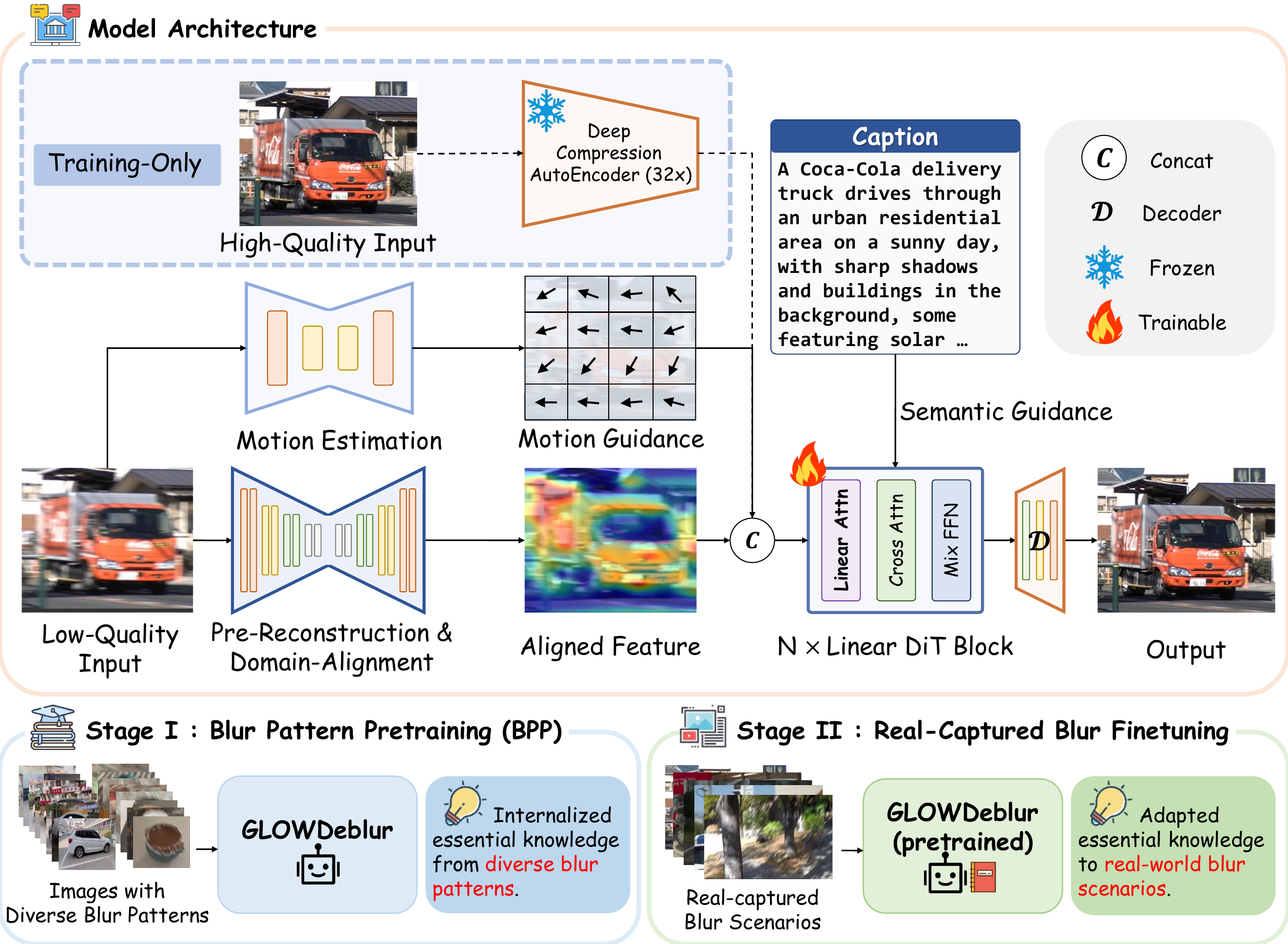}
    \vspace{-5pt}
    \caption{Overview of GLOWDeblur. The framework integrates a Pre-Reconstruction \& Domain-Alignment module with a lightweight diffusion framework, guided by motion maps and cross-modal text semantics. Training involves pre-training on datasets with diverse blur patterns, followed by joint fine-tuning on real-captured datasets.}
    \label{fig:method}
\end{figure*}

Given the demonstrated efficacy of BPP in bridging distribution gaps, we integrate this strategy into the training framework of GLOWDeblur. As illustrated in Figure.~\ref{fig:method}, the training process consists of two stages: the model first undergoes BPP on a simulated dataset featuring extensive blur coverage to internalize essential blur knowledge and priors. Subsequently, the model is fine-tuned on multiple real-world datasets, adapting these learned priors to realistic degradations to maximize both generalization capability and restoration quality.

\subsection{ Motion and Semantic Guidance (MoSeG) }
While BPP equips models with transferable blur priors, challenges remain under severe or highly diverse blur, where structural cues are ambiguous and low-level details are heavily lost. To address this, we introduce Motion and Semantic Guidance (MoSeG), a conditional design that explicitly reinforces blur priors during inference and training.

\textbf{Motion Guidance (MoG):} To strengthen the guidance of blur priors, we integrate a motion estimation module. Estimation of motion trajectories provides a direct way to characterize blur patterns and enhance the model’s ability to generalize across diverse degradations.  
The blur can be modeled as the accumulation of displaced sharp pixels along estimated trajectories: 
\begin{equation}
B(p_0) = \frac{1}{N} \sum_{n=0}^{N-1} L_s\!\left(p_0 + \Delta P_{t_n}\right),
\end{equation}
where $L_s$ is the latent sharp image and $\Delta P_{t_n}$ the motion offset at $t_n$. 


Following prior work on motion offset estimation~\cite{motion-estimate}, we adopt a lightweight encoder–decoder that extracts hierarchical features and predicts dense motion fields $\Delta P$. These offsets are concatenated with blurred-image features and fed into the deblurring network as motion cues.

\textbf{Semantic Guidance (SeG):} In severely blurred regions where structural details are lost, we inject high-level semantics as conditional signals to unleash the cross-modal capacity of diffusion models. Specifically, using Qwen2.5-VL-7B~\cite{Qwen2.5-VL}, we generate detailed captions describing objects, scenes, context, and other high-level attributes, and feed their embeddings into Linear DiT blocks, enabling the recovery of heavily degraded regions.

\begin{table*}[!b]
\centering
\caption{Quantitative comparison with state-of-the-art deblurring methods on six widely used benchmarks. Higher values indicate better performance for $\uparrow$ metrics, and lower values for $\downarrow$.}
\label{tab:results}
\scriptsize
\renewcommand{\arraystretch}{0.9}
\setlength{\tabcolsep}{3pt}
\begin{adjustbox}{center,scale=0.92}
\begin{tabular*}{0.85\textwidth}{@{\extracolsep{\fill}} c|c|cccccccc}
\toprule[1.5pt]
\textbf{Dataset} & \textbf{Metrics} & \textbf{Restormer*\cite{restormer}} & \textbf{HI-Diff\tiny \cite{hi-diff}} & \textbf{DiffIR\cite{diffir}} & \textbf{MISC-Filter\cite{miscfilter}} & \textbf{MLWNet\cite{mlwnet}} & \textbf{FPro\cite{fpro}} & \textbf{Diff-Plugin\cite{diff-plugin}} & \textbf{Ours} \\
\midrule
\multirow{10}{*}{\shortstack{GoPro \\ \cite{gopro}}} 
& PSNR $\uparrow$ & 33.07 & \textcolor{blue}{33.33}  & 33.20 & \textcolor{red}{34.10}  & 24.60 & 33.05 & 25.64 & 25.21 \\
& SSIM $\uparrow$ & 0.943 & \textcolor{blue}{0.964} & 0.963 & \textcolor{red}{0.969} & 0.83 & 0.943 & 0.793 & 0.787 \\
& MANIQA~\cite{maniqa} $\uparrow$ & 0.353 & 0.492 & \textcolor{blue}{0.535} & 0.458 & 0.497 & 0.518 & 0.346 & \textcolor{red}{0.538} \\
& LIQE~\cite{liqe} $\uparrow$ & 1.455 & 1.350 & \textcolor{red}{1.589} & 1.172 & 1.353 & 1.491 & 1.092 & \textcolor{blue}{1.502} \\
& NRQM~\cite{nrqm} $\uparrow$ & 4.748 & 5.047 & \textcolor{blue}{5.051} & 4.339 & 4.750 & 4.915 & 3.886 & \textcolor{red}{5.252} \\
& CLIP-IQA~\cite{clipiqa} $\uparrow$ & 0.243 & 0.250 & \textcolor{red}{0.258} & 0.214 & \textcolor{blue}{0.257} & 0.250 & 0.190 & 0.239 \\
& PI~\cite{pi} $\downarrow$ & 5.308 & 5.159 & \textcolor{blue}{5.135} & 5.464 & 5.363 & 5.151 & 6.202 & \textcolor{red}{4.846} \\
& BRISQUE~\cite{brisque} $\downarrow$ & 46.715 & 46.418 & 46.721 & \textcolor{blue}{46.095} & 49.638 & 49.121 & 51.018 & \textcolor{red}{39.732} \\
& NIQE~\cite{niqe} $\downarrow$ & 5.534 & 5.504 & 5.466 & 5.461 & 5.650 & \textcolor{blue}{5.377} & 6.146 & \textcolor{red}{5.120} \\
& ILNIQE~\cite{ilniqe} $\downarrow$ & 33.354 & 32.710 & 33.408 & 33.071 & \textcolor{blue}{32.535} & 32.701 & 42.474 & \textcolor{red}{26.464} \\
\midrule
\multirow{10}{*}{\shortstack{HIDE \\ \cite{hide}}}
& PSNR $\uparrow$ & \textcolor{red}{31.81} & 31.46 & 31.55 & \textcolor{blue}{31.66} & 23.95 & 30.70 & 23.95 & 24.12 \\
& SSIM $\uparrow$ & 0.933 & 0.945 & \textcolor{red}{0.947} & \textcolor{blue}{0.946} & 0.819 & 0.921 & 0.763 & 0.763 \\
& MANIQA $\uparrow$ & 0.453 & 0.535 & \textcolor{red}{0.592} & 0.498 & 0.509 & 0.572 & 0.362 & \textcolor{blue}{0.583} \\
& LIQE $\uparrow$ & 1.113 & 1.621 & \textcolor{red}{1.977} & 1.236 & 1.392 & \textcolor{blue}{1.803} & 1.061 & 1.788 \\
& NRQM $\uparrow$ & 3.916 & 5.613 & \textcolor{blue}{6.104} & 4.731 & 5.129 & 6.028 & 4.323 & \textcolor{red}{6.210} \\
& CLIP-IQA $\uparrow$ & 0.187 & \textcolor{red}{0.227} & \textcolor{red}{0.229} & 0.179 & 0.224 & 0.215 & 0.158 & 0.212 \\
& PI $\downarrow$ & 6.302 & 4.837 & 4.354 & 5.278 & 5.085 & \textcolor{red}{4.308} & 5.773 & \textcolor{red}{4.244} \\
& BRISQUE $\downarrow$ & 52.830 & 43.605 & 41.045 & 45.919 & 48.277 & 42.970 & \textcolor{blue}{40.716} & \textcolor{red}{36.687} \\
& NIQE $\downarrow$ & 6.558 & 5.254 & 4.803 & 5.318 & 5.348 & \textcolor{red}{4.624} & 5.528 & \textcolor{blue}{4.673} \\
& ILNIQE $\downarrow$ & 36.248 & 31.246 & 29.788 & 29.985 & 30.702 & \textcolor{blue}{28.224} & 40.744 & \textcolor{red}{24.176} \\
\midrule
\multirow{10}{*}{\shortstack{REDS  \\ \cite{reds}} }
& PSNR $\uparrow$ &  \textcolor{red}{34.207} & 25.760 & 26.78 & 27.58 & \textcolor{blue}{27.60} & 26.96 & 26.27 & 26.21 \\
& SSIM $\uparrow$ & \textcolor{red}{0.938} & 0.779 & 0.819 & 0.832 & \textcolor{blue}{0.851} & 0.840 & 0.771 & 0.770 \\
& MANIQA $\uparrow$ & 0.536 & 0.630 & 0.626 & 0.607 & \textcolor{red}{0.647} & 0.613 & 0.518 & \textcolor{blue}{0.642} \\
& LIQE $\uparrow$ & 1.435 & 2.530 & 2.293 & 2.065 & \textcolor{red}{2.664} & 2.097 & 1.520 & \textcolor{blue}{2.570} \\
& NRQM $\uparrow$ & 4.886 & 6.849 & \textcolor{blue}{7.014} & 6.541 & 6.954 & 6.809 & 6.384 & \textcolor{red}{7.352} \\
& CLIP-IQA $\uparrow$ & 0.271 & 0.305 & 0.287 & 0.268 & \textcolor{blue}{0.322} & 0.269 & 0.228 & \textcolor{red}{0.351} \\
& PI$\downarrow$ & 5.335 & 3.379 & 3.391 & 3.546 & \textcolor{blue}{3.293} & 3.481 & 4.160 & \textcolor{red}{3.035} \\
& BRISQUE $\downarrow$ & 43.364 & 28.061 & \textcolor{blue}{26.452} & 30.115 & 31.448 & 28.533 & 27.867 & \textcolor{red}{25.695} \\
& NIQE $\downarrow$ & 5.660 & 3.899 & 3.973 & 3.894 & \textcolor{blue}{3.851} & 3.966 & 4.620 & \textcolor{red}{3.694} \\
& ILNIQE $\downarrow$ & 29.305 & 23.784 & 23.269 & 23.083 & \textcolor{blue}{22.369} & 23.236 & 27.088 & \textcolor{red}{19.357} \\
\midrule
\multirow{10}{*}{\shortstack{RealBlur-J \\ \cite{realblur}}} 
& PSNR $\uparrow$ & 31.131 & 29.15 &  25.37 & \textcolor{red}{33.88} & \textcolor{blue}{33.84} &  27.90 & 26.25 & 27.55 \\
& SSIM $\uparrow$ & 0.917 & 0.890 & 0.825 & \textcolor{blue}{0.938} & \textcolor{red}{0.941} & 0.873 &  0.79 & 0.811 \\
& MANIQA$\uparrow$ & 0.472 & \textcolor{red}{0.629} & 0.571 & 0.602 & \textcolor{blue}{0.615} & 0.544 & 0.467 & 0.613 \\
& LIQE $\uparrow$ & 2.356 & \textcolor{red}{2.646} & 1.949 & 2.386 & \textcolor{blue}{2.578} & 1.735 & 1.243 & 2.439 \\
& NRQM $\uparrow$ & 5.150 & \textcolor{red}{5.870} & 5.517 & 5.365 & \textcolor{blue}{5.685} & 5.361 & 4.283 & 5.633 \\
& CLIP-IQA $\uparrow$ & 0.262 & \textcolor{red}{0.279} & 0.247 & 0.251 & \textcolor{blue}{0.274} & 0.212 & 0.208 & \textcolor{blue}{0.274} \\
& PI $\downarrow$ & 5.235 & \textcolor{red}{4.651} & 4.934 & 5.011 & 4.869 & 5.013 & 5.965 & \textcolor{blue}{4.787} \\
& BRISQUE $\downarrow$ & 49.636 & 46.799 & \textcolor{red}{40.207} & 46.610 & 48.970 & 42.742 & 42.401 & \textcolor{red}{35.895} \\
& NIQE $\downarrow$ & 5.708 & \textcolor{blue}{5.182} & 5.258 & 5.341 & 5.370 & 5.256 & 5.963 & \textcolor{red}{5.128} \\
& ILNIQE $\downarrow$ & 34.999 & 33.380 & \textcolor{blue}{31.848} & 34.550 & 33.588 & 32.580 & 37.317 & \textcolor{red}{27.548} \\
\midrule
\multirow{10}{*}{\shortstack{BSD \\ \cite{bsd}}} 
& PSNR $\uparrow$ & \textcolor{red}{30.410} & 28.66 & 27.97 & 29.53 & 28.82 & 26.64 & 27.67 & \textcolor{blue}{29.56} \\
& SSIM $\uparrow$ & \textcolor{red}{0.923} & 0.907 & 0.885 & \textcolor{red}{0.923} & \textcolor{blue}{0.910} & 0.885 & 0.862 & 0.893 \\
& MANIQA $\uparrow$ & \textcolor{red}{0.571} & 0.565 & 0.362 & 0.510 & 0.536 & 0.461 & 0.427 & \textcolor{blue}{0.568} \\
& LIQE $\uparrow$ & 2.325 & \textcolor{red}{2.452} & 1.048 & 1.742 & 2.132 & 1.479 & 1.296 & \textcolor{blue}{2.348} \\
& NRQM $\uparrow$ & 4.927 & \textcolor{red}{5.806} & 4.634 & 4.772 & \textcolor{blue}{5.229} & 5.141 & 4.433 & 5.096 \\
& CLIP-IQA $\uparrow$ & 0.279 & \textcolor{red}{0.283} & 0.188 & 0.234 & 0.264 & 0.179 & 0.195 & \textcolor{blue}{0.282} \\
& PI $\downarrow$ & 5.800 & \textcolor{red}{5.189} & 6.100 & 5.819 & \textcolor{blue}{5.560} & 5.934 & 6.422 & 5.589 \\
& BRISQUE $\downarrow$ & 47.363 & 39.518 & \textcolor{red}{24.113} & 46.358 & 46.388 & \textcolor{blue}{29.108} & 36.347 & 40.514 \\
& NIQE $\downarrow$ & 5.799 & \textcolor{blue}{5.464} & 6.469 & 5.758 & 5.721 & 5.881 & 6.569 & \textcolor{red}{5.455} \\
& ILNIQE $\downarrow$ & 42.040 & 38.708 & \textcolor{red}{31.513} & 41.328 & 40.506 & \textcolor{blue}{37.435} & 50.182 & 39.383 \\
\midrule
\multirow{10}{*}{\shortstack{RSBlur \\ \cite{rsblur}} }
& PSNR $\uparrow$ & 29.27 &  29.47 & 22.48 & \textcolor{blue}{29.98} & \textcolor{red}{30.91} &  26.19 & 27.82 & 28.85 \\
& SSIM $\uparrow$ & 0.864 & \textcolor{blue}{0.875} & 0.651 & \textcolor{red}{0.887} & 0.818 &  0.833 & 0.821 & 0.820 \\
& MANIQA $\uparrow$ & 0.442 & \textcolor{blue}{0.452} & 0.362 & 0.420 & 0.415 & 0.398 & 0.441 & \textcolor{red}{0.533} \\
& LIQE $\uparrow$ & \textcolor{blue}{1.342} & 1.124 & 1.048 & 1.069 & 1.111 & 1.018 & 1.015 & \textcolor{red}{1.404} \\
& NRQM $\uparrow$ & 3.769 & 3.817 & 4.634 & 3.523 & 3.642 & \textcolor{blue}{5.520} & 4.357 & \textcolor{red}{5.597} \\
& CLIP-IQA $\uparrow$ & \textcolor{red}{0.262} & 0.246 & 0.188 & 0.204 & \textcolor{blue}{0.248} & 0.170 & 0.169 & 0.236 \\
& PI $\downarrow$ & 6.427 & 6.851 & \textcolor{blue}{6.100} & 6.820 & 7.065 & 6.296 & 6.677 & \textcolor{red}{4.980} \\
& BRISQUE $\downarrow$ & 52.768 & 50.286 & \textcolor{blue}{24.113} & 54.119 & 58.433 & 39.250 & \textcolor{red}{21.942} & 30.677 \\
& NIQE $\downarrow$ & 6.522 & 7.348 & \textcolor{blue}{6.469} & 6.943 & 7.532 & 7.533 & 6.840 & \textcolor{red}{5.292} \\
& ILNIQE $\downarrow$ & 32.349 & 37.794 & \textcolor{blue}{31.513} & 36.354 & 39.651 & 36.035 & 41.705 & \textcolor{red}{25.833} \\
\bottomrule[1.5pt]
\end{tabular*}
\end{adjustbox}
\end{table*}

\begin{figure*}[!b]
    \centering
    \includegraphics[width=0.85\linewidth]{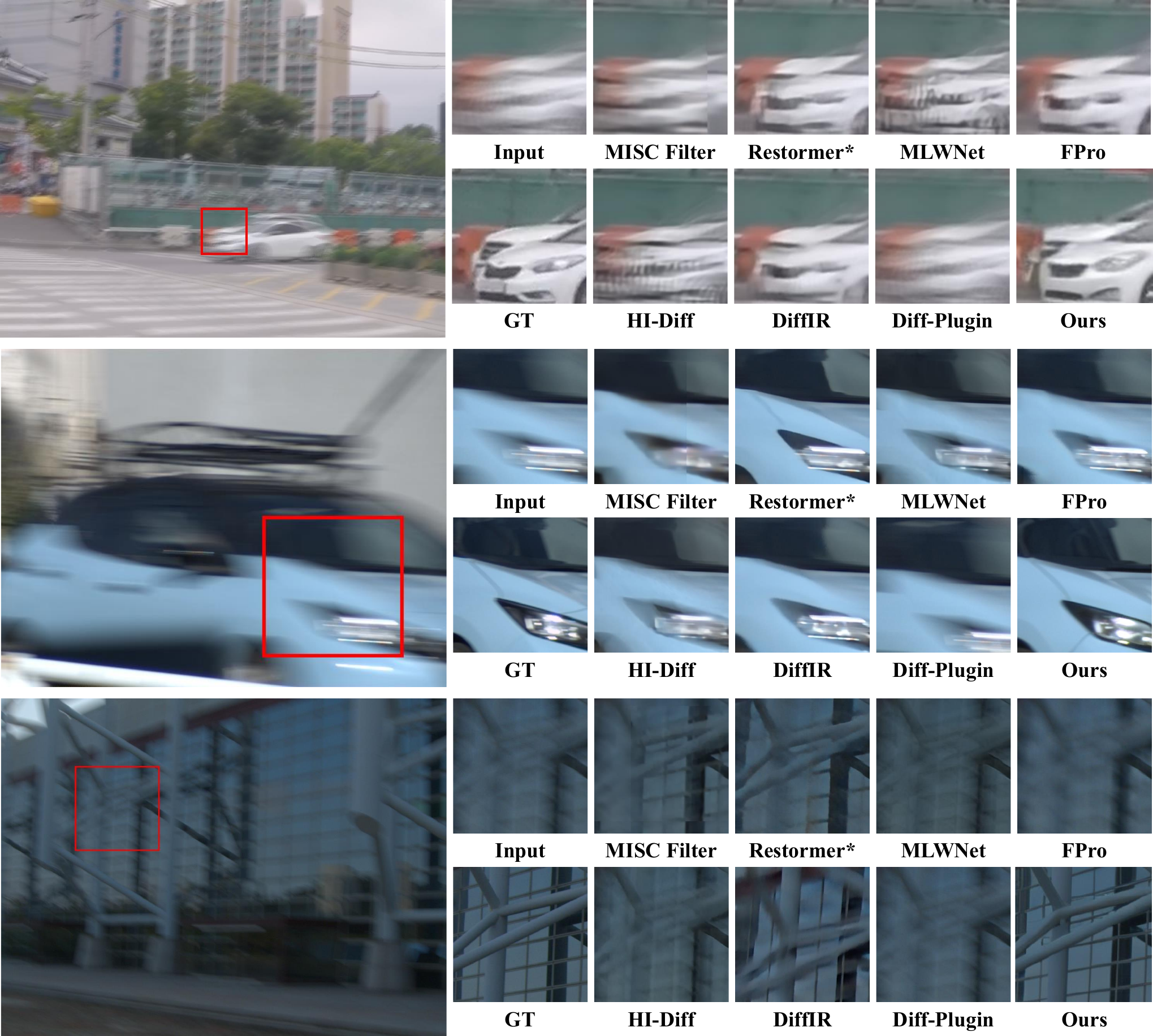}
    \caption{Qualitative comparison on  GoPro, BSD, and RSBlur (From top to bottom). GLOWDeblur effectively handles diverse blur patterns with high-quality restorations.}
    \label{fig:comparison_main_6datasets}
\end{figure*}

\subsection{Lightweight Pre-Aligned Linear Diffusion Framework}

Real-world deblurring applications, ranging from autonomous driving to mobile photography, demand models that are both highly efficient and compact. To this end, we design a lightweight framework that integrates a Pre-Reconstruction \& Domain-Alignment module with a Deep Compression AutoEncoder and Linear DiT blocks, achieving both efficiency and strong performance.

\textbf{Pre-Reconstruction \& Domain-Alignment module:} 
a conventional UNet architecture that provides coarse restoration and aligned intermediate representations,  reducing the burden on the diffusion backbone. To keep the design lightweight, we follow the philosophy of~\cite{naf}, simplifying architectures with two key modifications. First, nonlinear activations such as GELU are replaced with a SimpleGate, where feature maps are split and fused via element-wise product:
\begin{equation}
\text{SimpleGate}(X, Y) = X \odot Y,
\end{equation}
preserving gating capacity at negligible cost. Second, channel attention is reformulated as Simplified Channel Attention (SCA), which aggregates global context through pooled descriptors and reweights channels without redundant nonlinearities:

\begin{equation}
SCA(X) = X \odot W_{pool}(X).
\end{equation}
Together, these modifications substantially reduce computation while retaining representational power.

\textbf{Lightweight Diffusion with Deep Compression AutoEncoder and Linear Attention:}
Latent diffusion operates by compressing images into a latent space via an AutoEncoder and applying a DiT for diffusion within this space, where the computational cost is largely influenced by the compression ratio and the complexity of attention mechanisms. To meet real-world efficiency demands, we adopt lightweight adaptations inspired by ~\cite{sana,linearsr}.

While mainstream designs typically adopt an $8\times$ AutoEncoder for latent compression, we employ a more aggressive $32\times$ Deep Compression AutoEncoder. With sufficient performance maintained, this design reduces the number of tokens and significantly lowers memory and computation. 

Traditional Diffusion Transformers (DiTs) adopt the standard softmax attention mechanism with quadratic complexity $O(N^2)$, which is computationally expensive. we further replace the quadratic self-attention in the DiT with a linear variant, reducing the complexity to $O(N)$.

Given query $\mathbf{Q} \in \mathbb{R}^{N \times d}$, key $\mathbf{K} \in \mathbb{R}^{N \times d}$, and value $\mathbf{V} \in \mathbb{R}^{N \times d}$, the linear attention output is defined as:

\begin{equation}
\resizebox{0.6\columnwidth}{!}{$
\begin{aligned}
O_i &= \sum_{j=1}^N \frac{\mathrm{ReLU}(Q_i)\,\mathrm{ReLU}(K_j)^\top V_j}{\sum_{j=1}^N \mathrm{ReLU}(Q_i)\,\mathrm{ReLU}(K_j)^\top} \\
    &= \frac{\mathrm{ReLU}(Q_i)\Big(\sum_{j=1}^N \mathrm{ReLU}(K_j)^\top V_j\Big)}{\mathrm{ReLU}(Q_i)\Big(\sum_{j=1}^N \mathrm{ReLU}(K_j)^\top\Big)}
\end{aligned}
$}
\end{equation}

Instead of computing attention weights for every query--key pair, the shared terms $\sum_{j=1}^N \mathrm{ReLU}(K_j)^\top V_j \in \mathbb{R}^{d \times d} $ and  
$\sum_{j=1}^N \mathrm{ReLU}(K_j)^\top \in \mathbb{R}^{d \times 1}$
are computed only once, resulting in a lightweight and effective Linear DiT Block.

Finally, we adopt a lightweight fusion strategy that concatenates shallow convolutional features with motion guidance and latent representations before processing by the Linear DiT block, effectively integrating complementary cues while preserving efficiency.

\begin{table*}[!t]
\centering
\caption{Quantitative comparison with SOTA deblur models across real-world datasets RWBI and RWBlur400. Higher values are better for $\uparrow$ metrics, lower for $\downarrow$. Since these datasets lack ground-truth annotations, only no-reference metrics are reported.}
\label{tab:rwbi_webdata}
\renewcommand{\arraystretch}{1} 
\begin{adjustbox}{width=0.9\linewidth, center}
\begin{tabular}{c|c|ccccccccc}
\toprule[1.5pt]


\textbf{Dataset} & { \textbf{Metrics}} & \textbf{Restormer}\cite{restormer}& \textbf{Restormer*}\cite{restormer} & \textbf{HI-Diff}\cite{hi-diff} & \textbf{DiffIR}\cite{diffir} & \textbf{MISC-Filter}\cite{miscfilter} & \textbf{MLWNet}\cite{mlwnet} & \textbf{FPro}\cite{fpro} & \textbf{Diff-Plugin}\cite{diff-plugin} & \textbf{Ours} \\

\midrule
\multirow{8}{*}{\shortstack{RWBI \\ \cite{rwbi}}} 
& MANIQA~\cite{maniqa} $\uparrow$     & 0.501 & 0.522 & 0.554 & 0.494 & 0.492 & \textcolor{blue}{0.565} & 0.483 & 0.460 & \textcolor{red}{0.635} \\
& LIQE~\cite{liqe} $\uparrow$       & 1.846 & 2.180 & 2.875 & 1.807 & 2.150 & \textcolor{blue}{3.068} & 1.771 & 1.371 & \textcolor{red}{3.732} \\
& NRQM~\cite{nrqm} $\uparrow$       & 5.433 & 5.608 & 5.879 & 5.457 & 5.079 & \textcolor{blue}{6.185} & 5.474 & 5.028 & \textcolor{red}{6.393} \\
& CLIP-IQA~\cite{clipiqa} $\uparrow$    & 0.257 & 0.283 & 0.372 & 0.256 & 0.301 & \textcolor{blue}{0.424} & 0.236 & 0.234 & \textcolor{red}{0.474} \\
& PI~\cite{pi} $\downarrow$       & 5.291 &5.133 & 4.993 & 5.353 & 5.468 & \textcolor{red}{4.459} & 5.182 & 5.670 & \textcolor{blue}{4.789} \\
& BRISQUE~\cite{brisque} $\downarrow$  & 39.945 & 39.192 & 40.674 & 42.319 & 43.271 & \textcolor{blue}{37.403} & 39.581 & 40.488 & \textcolor{red}{36.625} \\
& NIQE~\cite{niqe} $\downarrow$     & 5.682 &5.545 & 5.589 & 5.831 & 5.793 & \textcolor{blue}{4.886} & 5.550 & 5.988 & \textcolor{red}{4.796} \\
& ILNIQE~\cite{ilniqe} $\downarrow$   & 40.760 & 37.569 & 37.047 & 41.304 & 38.705 & \textcolor{blue}{34.259} & 40.208 & 47.048 & \textcolor{red}{33.271} \\
\midrule
\multirow{8}{*}{RWBlur400} 
& MANIQA $\uparrow$    & 0.509& \textcolor{blue}{0.530} & 0.490 & 0.493 & 0.455 & 0.517 & 0.481 & 0.497 & \textcolor{red}{0.604} \\
& LIQE $\uparrow$       & 1.746 & 1.893 & 2.041 & 1.983 & 1.780 & \textcolor{blue}{2.136} & 1.844 & 1.808 & \textcolor{red}{2.390} \\
& NRQM $\uparrow$       & 4.956 & 5.471 & 5.411 & 5.747 & 4.901 & 5.736 & \textcolor{blue}{5.783} & 5.660 & \textcolor{red}{6.746} \\
& CLIP-IQA $\uparrow$    & 0.333 & 0.352 & \textcolor{blue}{0.367} & 0.339 & 0.305 & 0.362 & 0.313 & 0.360 & \textcolor{red}{0.459} \\
& PI $\downarrow$       & 4.788 & 4.734 & 4.990 & 4.578 & 5.236 & \textcolor{blue}{4.461} & 4.520 & 4.649 & \textcolor{red}{3.786} \\
& BRISQUE $\downarrow$  & 41.083 & 40.620 & 43.704 & 34.578 & 46.365 & 41.043 & 31.025 & \textcolor{blue}{30.259} & \textcolor{red}{27.956} \\
& NIQE $\downarrow$     & 4.971 & 4.817 & 5.229 & 4.751 & 5.204 & \textcolor{red}{4.575} & 4.609 & 4.777 & \textcolor{blue}{4.576} \\
& ILNIQE $\downarrow$   & \textcolor{blue}{31.732} & 31.796 & 34.968 & 32.232 & 34.699 & 32.491 & 32.749 & 33.989 & \textcolor{red}{29.809} \\
\bottomrule[1.5pt]
\end{tabular}
\end{adjustbox}
\end{table*}

\begin{figure*}[!t]
    \centering
    \includegraphics[width=0.7\linewidth]{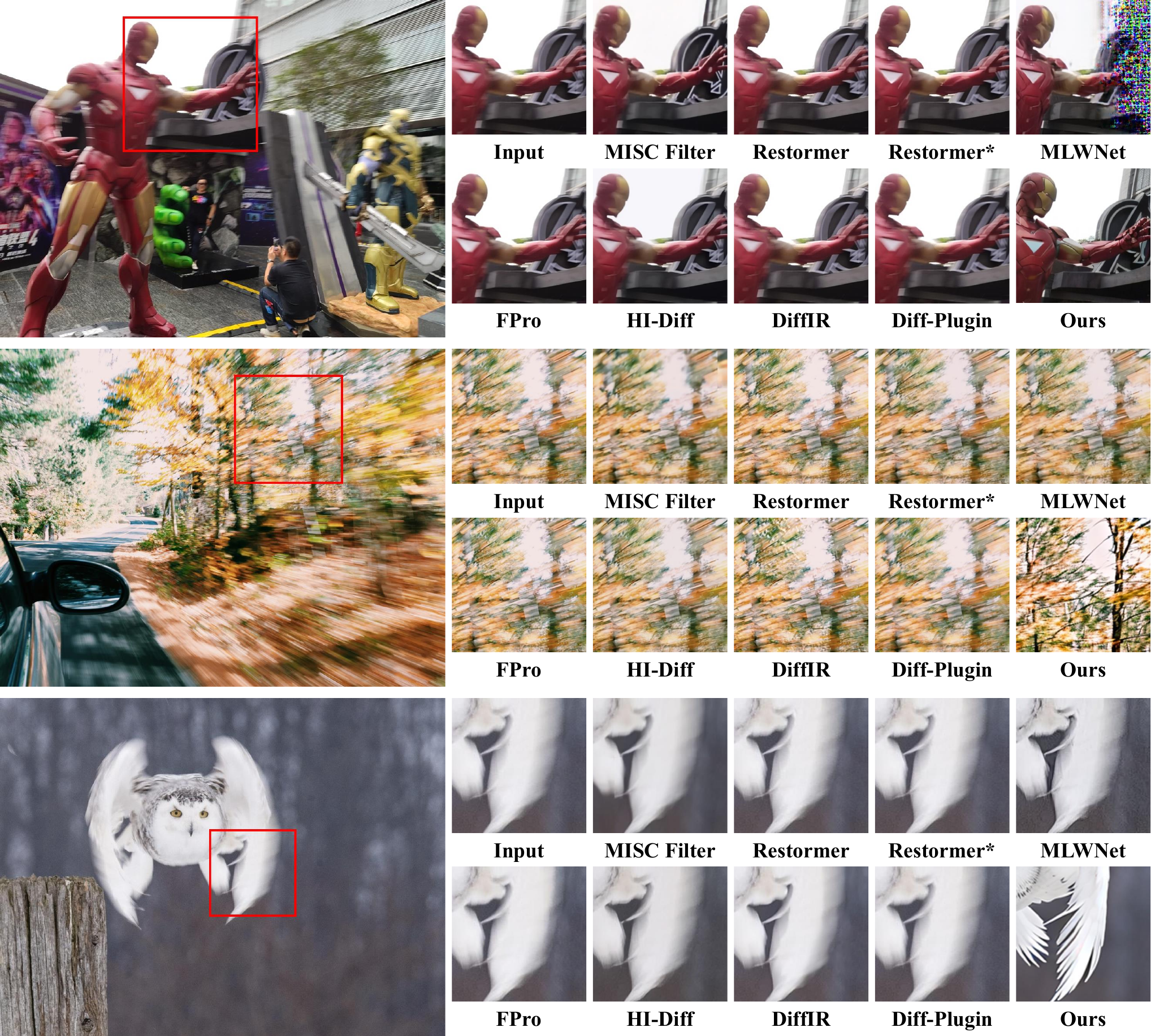}
    \caption{Comparison with SOTA deblur models on real-world datasets RWBI and RWBlur400.}
    \label{fig:comparison_main_real}
\end{figure*}

\section{Experiments}
\subsection{Experiment Settings}

Training proceeds in two stages. First, BPP is performed on simulated datasets, including GSBlur (3D Gaussian Splatting with randomized camera trajectories) and an augmented subset of LSDIR~\cite{lsdir} (by simply adding Gaussian and motion blur at different levels). Although less realistic, these datasets provide broad blur pattern coverage . Second, we jointly fine-tune on GoPro, HIDE, REDS, RealBlur, BSD, and RSBlur to align these priors with real-world distributions and enhance restoration quality. More implementation details are given in the Appendix Sec.~\ref{sec:appendix_implementation_details}.

We evaluate methods on six real-captured datasets for cross-dataset generalization. For real-world applicability, we further test on RWBI~\cite{rwbi} and our collected RWBlur400. Distinguished from existing benchmarks that are limited to street views, RWBlur400 is widely collected from the web, providing comprehensive coverage of diverse scene categories and complex blur patterns. Visual examples and detailed descriptions are provided in Appendix Sec.~\ref{sec:appendix_RWBlur400_details}.
Overall, we employ both reference-based and no-reference metrics to comprehensively evaluate deblurring performance. For reference-based evaluation, we use PSNR and SSIM. However, since recent image restoration research increasingly prioritizes perceptual quality over purely pixel-wise fidelity, our evaluation also places a strong emphasis on perceptual aspects. To this end, we adopt a diverse set of no-reference quality metrics—including MANIQA~\cite{maniqa}, LIQE~\cite{liqe}, NRQM~\cite{nrqm}, CLIP-IQA~\cite{clipiqa}, PI~\cite{pi}, BRISQUE~\cite{brisque}, NIQE~\cite{niqe}, and ILNIQE~\cite{ilniqe}, which together provide a thorough assessment of perceptual quality and better reflect real-world visual performance.

\subsection{Comparisons with State of the Arts}

We compare GLOWDeblur with state-of-the-art approaches from two categories. The first includes recent deblurring-specific methods such as HI-Diff, MISCFilter, and MLWNet. The second covers general restoration frameworks such as Restormer (and Restormer* retrained under our pipeline), DiffIR, FPro, and Diff-Plugin. Both categories contain a mix of diffusion-based and non-diffusion baselines, allowing a fair and comprehensive evaluation.

\subsubsection{Cross-Dataset Generalization}

We evaluate cross-dataset generalization on six widely used datasets. As shown in Figure.~\ref{tab:results}, GLOWDeblur mitigates cross-dataset distribution gaps, achieving strong deblurring performance and high-quality restoration with competitive fidelity. Restormer* also achieves good fidelity across different datasets. Figure.\ref{fig:comparison_main_6datasets} provides qualitative comparisons, with additional results presented in Appendix Sec.\ref{sec:VisualResults}. These results demonstrate that GLOWDeblur robustly handles complex blur patterns and produces high-quality restorations. More results are provided in the Appendix Sec.~\ref{sec:VisualResults}.
\begin{table*}[!t]
\centering
\caption{Ablation studies on REDS and RSBlur. \colorbox{gray!20}{Gray} indicates the settings of GLOWDeblur.}
\label{tab:ablation}
\renewcommand{\arraystretch}{1.1} 
\setlength{\tabcolsep}{6pt}      

\begin{adjustbox}{width=0.88\linewidth, center}
\begin{tabular}{@{}cc ccc cc cccc@{}} 
\toprule[1.5pt]

\multirow{2}{*}{\textbf{Dataset}} & \multirow{2}{*}{\textbf{No.}} &
\multicolumn{3}{c}{\textbf{Model Modules}} &
\multicolumn{2}{c}{\textbf{Data Pipeline}} &
\multicolumn{4}{c}{\textbf{Metrics}} \\

\cmidrule(lr){3-5} \cmidrule(lr){6-7} \cmidrule(lr){8-11}

& &
$N_{\text{Pre-aline}}$ &
$G_{\text{motion}}$ &
$G_{\text{text}}$ &
BPP & Mix all &
MANIQA~\cite{maniqa} $\uparrow$ &
LIQE~\cite{liqe} $\uparrow$ &
NRQM~\cite{nrqm} $\uparrow$ &
BRISQUE~\cite{brisque} $\downarrow$ \\

\midrule

\multirow{5}{*}{REDS~\cite{reds}} & (a) & \xmark & \xmark & \xmark & \cmark & \xmark & 0.565 & 1.817 & 5.968 & 34.297 \\
& (b) & \cmark & \xmark & \xmark & \cmark & \xmark & 0.605 & 2.099 & 6.250 & 32.230 \\
& (c) & \cmark & \cmark & \xmark & \cmark & \xmark & 0.635 & 2.480 & 7.340 & 27.810 \\
\rowcolor{gray!20} \cellcolor{white}
& (d) & \cmark & \cmark & \cmark & \cmark & \xmark & \textbf{0.642} & \textbf{2.570} & \textbf{7.352} & \textbf{25.695} \\
& (e) & \cmark & \cmark & \cmark & \xmark & \cmark & 0.608 & 2.255 & 6.014 & 31.794 \\
\midrule

\multirow{5}{*}{RSBlur~\cite{rsblur}} & (a) & \xmark & \xmark & \xmark & \cmark & \xmark & 0.463 & 1.132 & 4.392 & 43.854 \\
& (b) & \cmark & \xmark & \xmark & \cmark & \xmark & 0.484 & 1.181 & 4.715 & 39.940 \\
& (c) & \cmark & \cmark & \xmark & \cmark & \xmark & 0.526 & 1.369 & 5.421 & 32.578 \\
\rowcolor{gray!20}\cellcolor{white}
& (d) & \cmark & \cmark & \cmark & \cmark & \xmark & \textbf{0.533} & \textbf{1.404} & \textbf{5.597} & \textbf{30.677} \\
& (e) & \cmark & \cmark & \cmark & \xmark & \cmark & 0.495 & 1.283 & 5.057 & 41.510 \\
\bottomrule[1.5pt]

\end{tabular}
\end{adjustbox}
\end{table*}

\subsubsection{Real-World Evaluation}

We evaluate real-world performance on RWBI and collected RWBlur400 datasets. As shown in Table.~\ref{tab:rwbi_webdata}, GLOWDeblur consistently outperforms state-of-the-art baselines, demonstrating superior performance under real-world degradations. Restormer* also achieves significant improvements compared to its original version. Figure.~\ref{fig:comparison_main_real} presents qualitative comparisons, where existing methods fail under severe blur, but GLOWDeblur produces clear and reliable restorations in real-world scenarios. More results are provided in the Appendix Sec.~\ref{sec:VisualResults}.

\subsection{Ablation Studies}

We conduct ablation studies on REDS and RSBlur (Table.~\ref{tab:ablation}). The Pre-Reconstruction \& Domain-Alignment module provides consistent gains (a,b) by stabilizing representations and easing the diffusion backbone. Motion guidance enhances blur priors with trajectory cues (b,c), while semantic guidance introduces high-level semantics to recover severely degraded regions (c,d). Replacing BPP with naïve mixed-data training causes clear drops (d,e), confirming its role in bridging cross-dataset gaps. Overall, these results validate the effectiveness of both BPP and the model components. Qualitative results of motion guidance and semantic guidance are provided in the Appendix Sec.~\ref{sec:SeG_and_moG}.

\section{Conclusion}
\label{sec:Conclusion}

In this work, we identify blur pattern diversity as key to generalization and propose GLOWDeblur, a lightweight diffusion-based framework that integrates Blur Pattern Pretraining (BPP) and Motion \& Semantic Guidance (MoSeG), achieving state-of-the-art performance on multiple synthetic and real-world datasets with substantially stronger generalization.

\clearpage

{
    \small

}

\title{
    Toward Generalizable Deblurring: Leveraging Massive Blur Priors with Linear Attention for Real-World Scenarios
}

\clearpage

\setcounter{page}{1}
\maketitlesupplementary

\section{Related Works}

\subsection{Image Deblurring}

Image deblurring has long been a fundamental problem in low-level vision. Earlier methods mainly relied on handcrafted priors and optimization-based formulations, such as gradient sparsity, edge sharpness, or statistical constraints~\cite{related—earlymethod1,related—earlymethod2}. While these approaches provided valuable insights, their strong reliance on manually designed assumptions made them inadequate for handling complex and diverse real-world blur. With the rise of deep learning, researchers have shifted toward data-driven architectures, enabling significant improvements in both restoration quality and efficiency. In this work, we focus on the latest generation of learning-based approaches that have recently achieved state-of-the-art performance in deblurring and general image restoration.

Task-specific deblurring architectures~\cite{stripformer,miscfilter,mlwnet,hi-diff}  have been extensively explored. Non-diffusion approachesintroduce specialized designs tailored for motion blur removal. ~\cite{stripformer} employs directional strip-based attention to capture region-specific blur orientations and magnitudes efficiently. ~\cite{miscfilter} leverages motion-adaptive collaborative filtering to handle spatially variant motion in real-world settings. ~\cite{mlwnet} integrates multi-scale prediction with learnable wavelet transforms to preserve frequency and directional continuity. On the diffusion side, ~\cite{hi-diff} designs a compact latent diffusion model with hierarchical integration to generate blur-aware priors for regression-based restoration. These specialized models typically achieve strong performance in blur removal but often generalize poorly when facing unseen blur patterns.

Beyond specialized models, general-purpose restoration frameworks~\cite{uformer,restormer,diffir,diff-plugin,fpro} have also been widely applied to deblurring. Non-diffusion methods demonstrate strong versatility across tasks. ~\cite{uformer} employs locally enhanced window attention to scale to high-resolution restoration,~\cite{restormer} introduces channel-wise self-attention for efficient global context modeling , and ~\cite{fpro} incorporates frequency prompting to guide restoration across different degradations. Diffusion-based methods~\cite{diffir,diff-plugin} further extend general restoration: ~\cite{diffir} integrates compact priors into efficient denoising diffusion, and ~\cite{diff-plugin} introduces lightweight task-specific plugin modules to adapt pre-trained diffusion models across diverse low-level vision tasks. Compared with task-specific designs, these general frameworks exhibit stronger cross-task robustness and generalization, though they often lag behind specialized models in task-optimized fidelity.

Despite these advances, most existing approaches still struggle with generalization in real-world scenarios. While task-specific methods achieve strong performance under their training distributions, they often fail to transfer across diverse blur patterns. General-purpose frameworks, though more robust across degradations, tend to sacrifice task-optimized fidelity. Overall, systematic investigation into real-world generalization for deblurring remains limited, leaving a critical gap that our work aims to address.

\subsection{Deblurring Datasets}
Progress in image deblurring has been closely tied to the availability of datasets. Yet, constructing suitable datasets is inherently challenging, as it involves balancing realism, diversity, and scalability. Synthetic datasets~\cite{gopro,hide,reds} such as GoPro~\cite{gopro}, HIDE~\cite{hide}, and REDS~\cite{reds} have been the dominant benchmarks for years. They are generated through pipelines that average or interpolate high-frame-rate videos to simulate camera exposure, offering large-scale paired data at relatively low cost. Such datasets have enabled rapid progress by providing standardized benchmarks, but the blur they simulate often deviates from real imaging processes. As a result, models trained on synthetic data may perform well in-distribution but fail to capture the irregular, spatially variant blur patterns observed in the wild.

To reduce this gap, real-captured datasets~\cite{realblur,rsblur,bsd} have been developed. RealBlur~\cite{realblur}, BSD~\cite{bsd}, and RSBlur~\cite{rsblur} adopt specialized imaging systems—such as beam-splitter setups or synchronized multi-camera rigs—to capture geometrically aligned pairs of blurred and sharp images. These datasets provide authentic motion and defocus blur, more faithfully reflecting the complexity of real-world degradations. However, the hardware cost and collection complexity are significant, limiting the dataset scale and diversity. Even with substantial effort, it remains nearly impossible to comprehensively cover the range of blur magnitudes, orientations, and scene dynamics encountered in real scenarios.

Recently, simulation-based datasets such as GSBlur~\cite{gsblur} have been proposed to improve diversity and controllability. By reconstructing 3D scenes with Gaussian Splatting and rendering them under randomized camera trajectories, GSBlur generates blur patterns beyond traditional frame-averaging pipelines. While this controllability broadens the degradation space, simulated blur still lacks the photometric and structural fidelity of real imaging, leaving a clear gap to real-captured datasets.

In summary, synthetic datasets are abundant but unrealistic, real-captured ones are authentic but costly and narrow, and simulation-based ones offer diversity but lack realism. No dataset achieves both scale and fidelity, creating distribution gaps that cause models to overfit specific blur patterns and degrade sharply under unseen conditions. This underscores the need for strategies that explicitly address blur diversity and distribution mismatch, motivating our work.

\subsection{Diffusion Models}
Diffusion Models (DMs)~\cite{diffusion1,diffusion2,diffusion3} have recently emerged as powerful generative priors, synthesizing data from Gaussian noise through iterative denoising. Their success in image generation has inspired a series of applications in  deblurring. In the context of deblurring, DiffIR~\cite{diffir} and HI-Diff~\cite{hi-diff} adopt diffusion-based priors with a two-stage training strategy to better capture blur statistics, More recently, IDBlau~\cite{idblau}leverages implicit diffusion to augment blur patterns under controllable settings, effectively enriching training data for downstream deblurring models.

Despite their effectiveness, most of these approaches remain computationally expensive. Large-scale pretrained diffusion models ~\cite{supir,diffusion1,dreamclear}possess billions of parameters, which, while offering strong generative priors, impose prohibitive training and inference costs that limit deployment in real-world scenarios like autonomous driving and mobile imaging. This challenge has motivated efforts to develop lightweight alternatives. For example, ~\cite{sana} proposes a linear-attention-based diffusion transformer that achieves high efficiency without sacrificing quality, demonstrating that architectural re-design and aggressive compression can bring diffusion models closer to practical deployment. Similarly, ~\cite{linearsr} and related works explore simplified diffusion formulations tailored for image restoration. Nonetheless, the exploration of lightweight diffusion for deblurring remains limited, leaving an open question of how to balance generalization, restoration fidelity, and efficiency under real-world constraints.

\section{Implementation Details}
\label{sec:appendix_implementation_details}
Our model is trained in two stages. First, Blur Pattern Pretraining (BPP) is performed for 10k iterations on a synthetic mixture of GSBlur and an augmented subset of LSDIR, where Gaussian and motion blur of varying levels are added to enrich pattern diversity. The model is then fine-tuned until convergence on a combined real-captured dataset including GoPro, HIDE, REDS, RealBlur, BSD, and RSBlur, aligning the learned priors with real-world distributions. Training is conducted using the Adam optimizer with an initial learning rate of 1e-4 and a batch size of 12×8. All experiments are implemented in PyTorch and run on 8 NVIDIA A800 GPUs (80GB each).

\section{Comprehensive Efficiency Gains from Our Lightweight Improvements}

To meet the stringent efficiency demands of real-world deblurring deployment, our lightweight Pre-Aligned Linear Diffusion framework is specifically designed to address the common limitations of classical diffusion models such as large parameter counts and slow inference, as reflected in SDXL and FLUX-dev. By employing a 32× Deep Compression AutoEncoder to substantially reduce the number of latent tokens, and adopting a Linear DiT that replaces quadratic attention with an O(N) linear variant, our approach effectively removes the major computational bottlenecks found in existing diffusion systems. This lightweight design enables near-linear scaling on high-resolution inputs, which is crucial for real-time deblurring scenarios including mobile photography, online services, and embedded vision applications. 

As shown in Table.~\ref{tab:comparison}, our method achieves the highest throughput, the lowest latency, and the smallest parameter footprint, demonstrating that the linear-attention-based architecture provides a practical and highly efficient alternative to traditional diffusion frameworks.

\begin{table}[h]
\centering
\caption{Comparison of Throughput, Latency, and Model Parameters}
\resizebox{1\columnwidth}{!}{%
\begin{tabular}{lccc}
\toprule
\textbf{Methods} & \textbf{Throughput ($\uparrow$)} & \textbf{Latency ($\downarrow$)} & \textbf{Params ($\downarrow$)} \\
                 & (Samples/s) & (s/Sample) & (B) \\
\midrule
FLUX-dev         & 0.09 & 11.0 & 12.0 \\
SDXL             & 0.3 & 3.3 & 2.6 \\
Ours             & 1.7 & 0.6 & 1.6 \\
\bottomrule
\end{tabular}
}

\label{tab:comparison}
\end{table}

\section{Qualitative Results of Motion Guidance and Semantic Guidance}
\label{sec:SeG_and_moG}
\begin{figure}[H]
    \centering
    \includegraphics[width=0.7\linewidth]{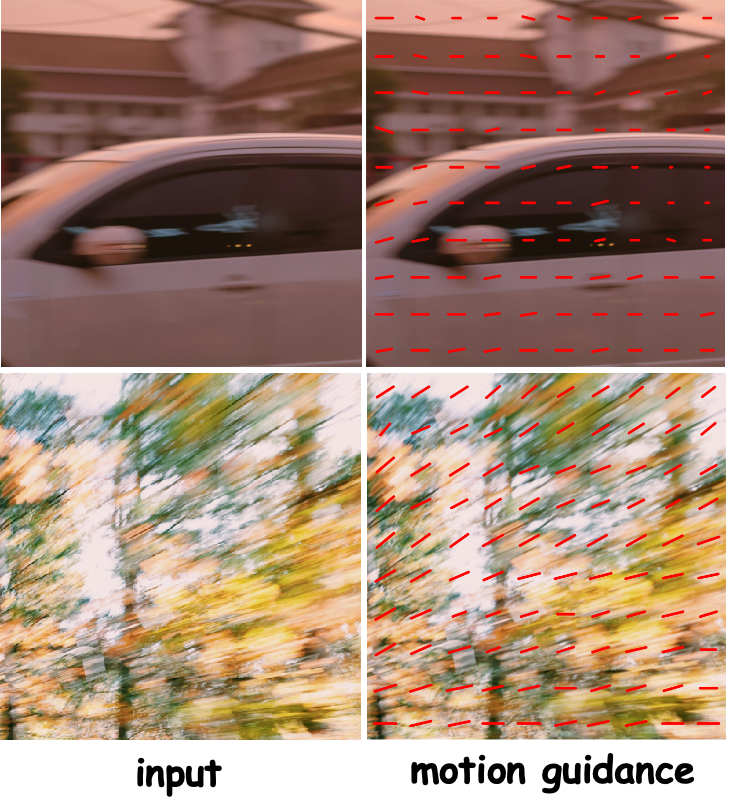}
    \caption{Qualitative results of motion guidance. }
    \label{fig:ablation_motion}
\end{figure}

To visualize the effectiveness of our motion–aware design, we divide each input image into a 10×10 grid of patches and estimate the dominant motion direction within each patch. The aggregated patch-wise motion vectors are then rendered as line segments over the image, providing an intuitive qualitative representation of the underlying blur field. As shown in~\ref{fig:ablation_motion}, our motion guidance accurately captures both the direction and the approximate magnitude of the local blur, even in scenes with spatially varying or non-uniform motion. The recovered motion structure exhibits a strong correspondence to the true blur patterns, revealing clear, coherent streak orientations that align with the motion that produced the blur.

Such explicit motion cues offer the model reliable, localized information about how the image was degraded. This strengthens the role of blur pattern awareness in the restoration process and allows the network to better adapt to complex motion distributions, ultimately enabling more faithful recovery compared with variants lacking motion guidance.

\begin{figure}[H]
    \centering
    \includegraphics[width=1.0\linewidth]{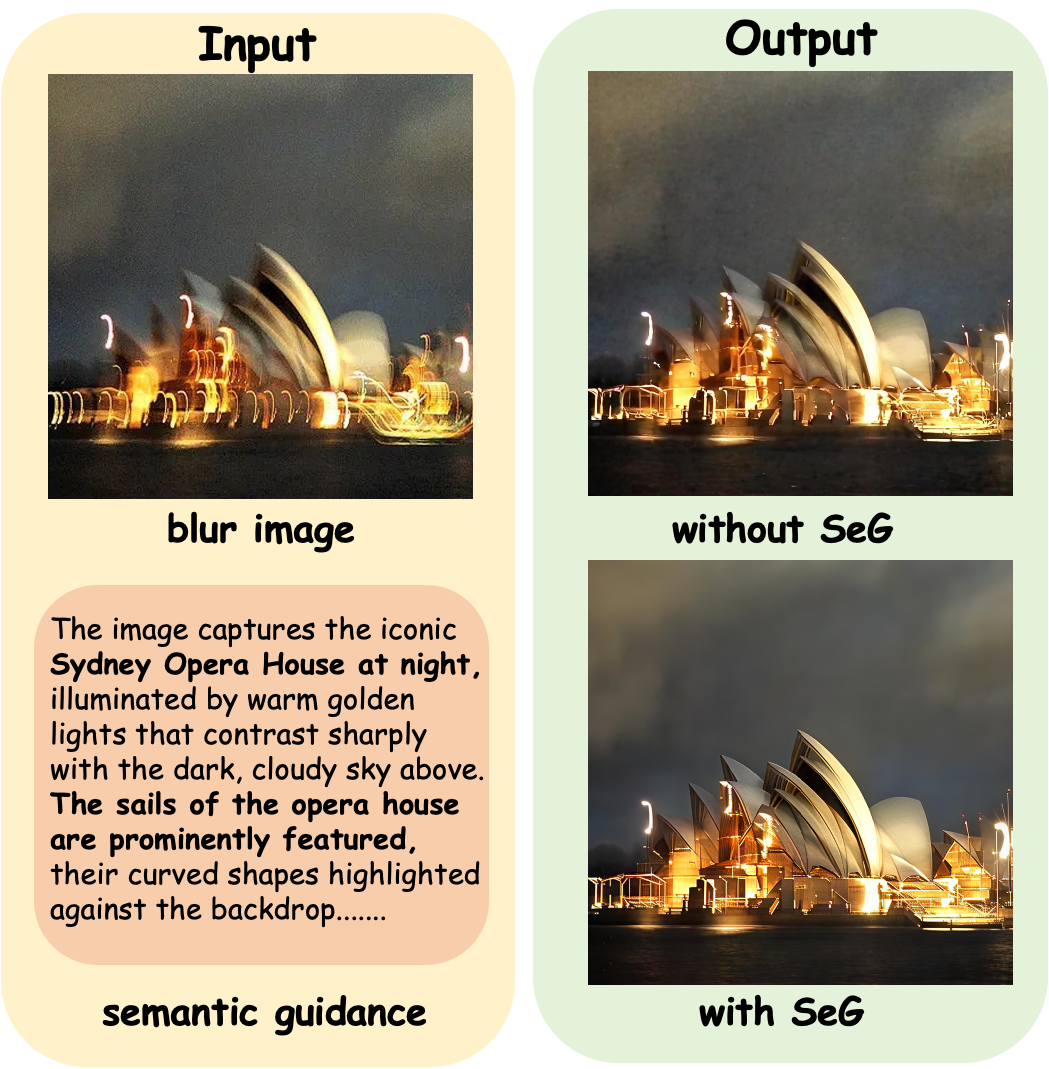}
    \caption{Qualitative results of semantic guidance. }
    \label{fig:ablation_semantic}
\end{figure}

Beyond low-level motion cues, our semantic guidance (SeG) introduces high-level structural priors through text captions, which is particularly beneficial when restoring regions that suffer from severe degradation. As illustrated in Fig.~\ref{fig:ablation_semantic}, when the input image is heavily blurred, the visual evidence alone provides insufficient information for accurate reconstruction. In contrast, the semantic description—containing explicit references such as “Sydney Opera House at night” and the “prominent sail-shaped roofs”—offers strong contextual constraints that guide the model toward plausible structural recovery. These semantic cues enable the network to better infer the global architecture and restore fine details that would otherwise be irretrievable. Consequently, SeG substantially enhances the model’s ability to handle severely blurred regions, yielding results with significantly improved clarity .

\begin{figure*}[!t]
    \centering
    \includegraphics[width=0.7\linewidth]{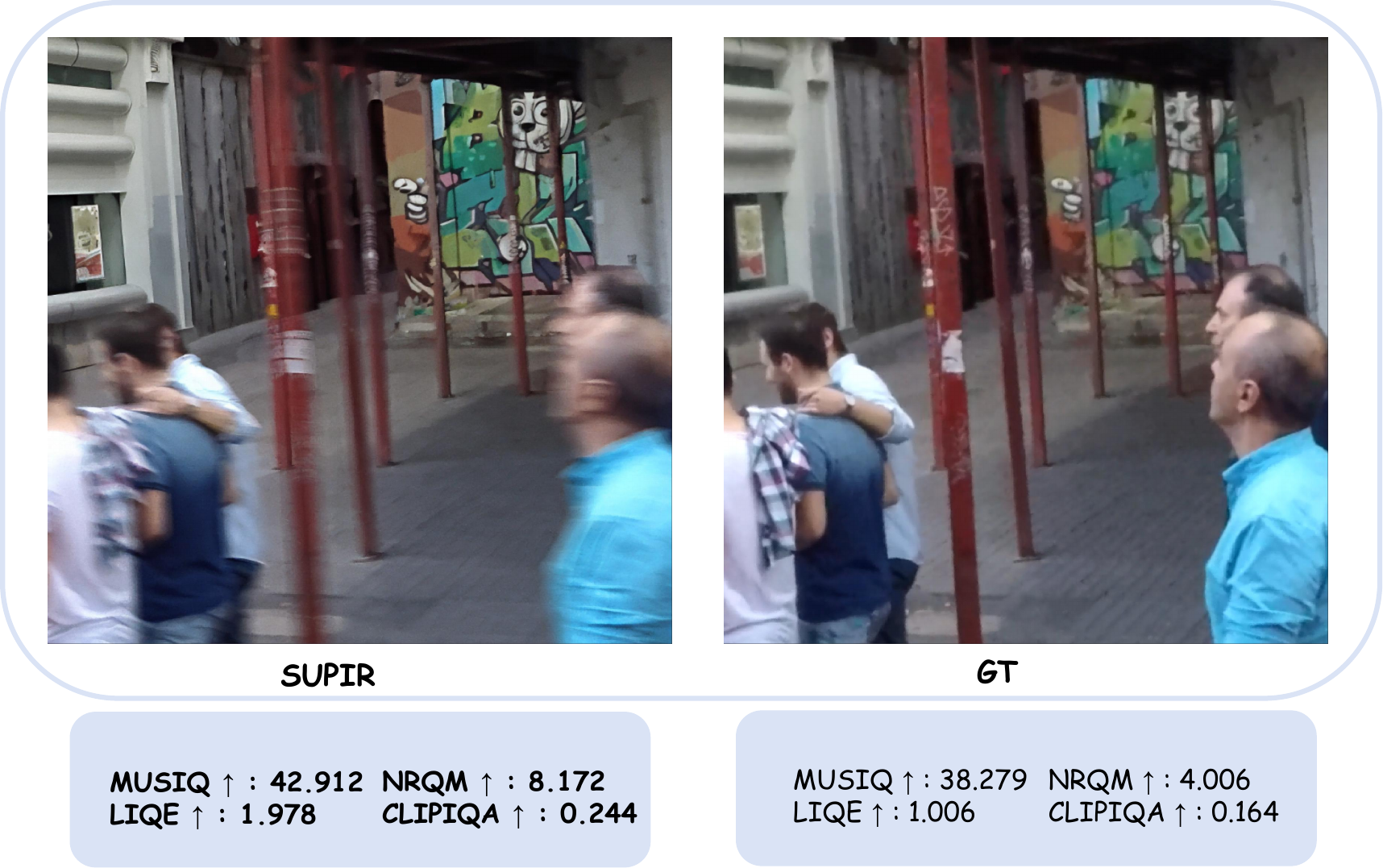}
    \caption{Qualitative and quantitative comparison of SUPIR and GT on GoPro, showing quality scores exceeding GT but failure to remove blur. }
    \label{fig:supir}
\end{figure*}

\section{Exploratory Comparison with Large-Scale General-Purpose Restoration Models}

With the rapid development of diffusion models, large-scale variants trained on massive datasets have demonstrated impressive capabilities across diverse image restoration tasks~\cite{diffusion1,supir,dreamclear}. These models, equipped with hundreds of millions or even billions of parameters, form the backbone of several general-purpose restoration frameworks that achieve state-of-the-art results in super-resolution, denoising, and image quality enhancement. Motivated by their success, we further investigate whether such models can leverage their scale and training data to generalize to real-world deblurring.

\begin{figure*}[!t]
    \centering
    \includegraphics[width=0.7\linewidth]{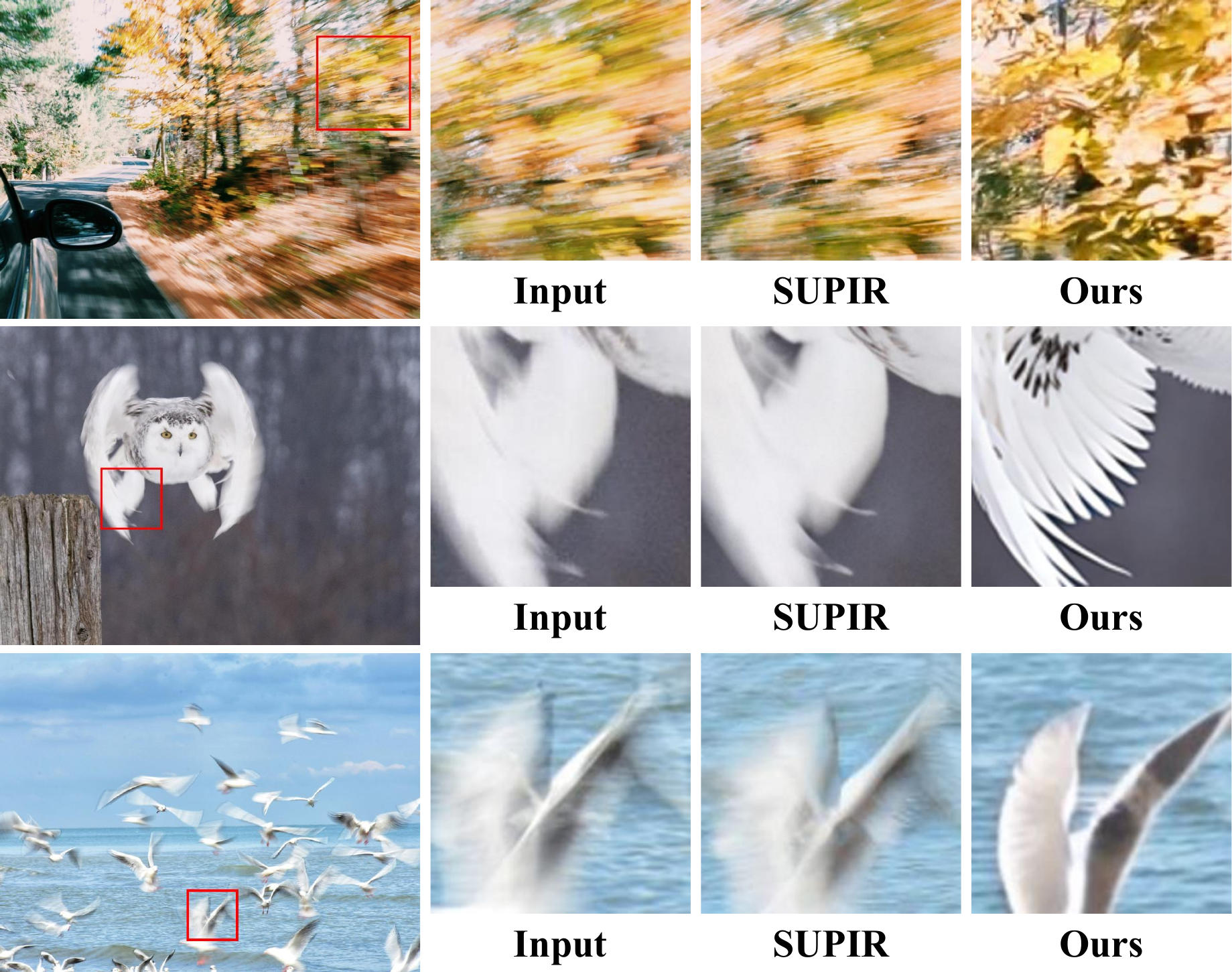}
    \caption{Quantitative comparison with SUPIR  across real-world datasets. }
    \label{fig:supir_comp}
\end{figure*}

However, our experiments reveal notable limitations. Using SUPIR~\cite{supir}as a representative model, we find that while it excels in enhancing perceptual quality—sometimes even surpassing ground-truth images in conventional quality metrics—it fails to effectively handle blur. As shown in ~\ref{fig:supir}, SUPIR struggles even on GoPro, one of the simplest synthetic benchmarks for motion deblurring, producing visually sharp but still blurred outputs. More strikingly, Figure.~\ref{fig:supir_comp} also illustrates its shortcomings on complex real-world blur, where artifacts and residual degradation remain prominent.

These observations highlight an important gap: despite their remarkable success in other restoration tasks, large-scale diffusion models are not inherently equipped to handle the structural complexity of blur. This contrast further validates the necessity of explicitly modeling blur priors, as pursued in our proposed framework, to achieve robust and generalizable deblurring in real-world scenarios.


\twocolumn[{%
    
    \section{Additional Visual Results}
    \label{sec:VisualResults}
    \vspace{5pt}

    \centering
    \includegraphics[width=0.95\linewidth]{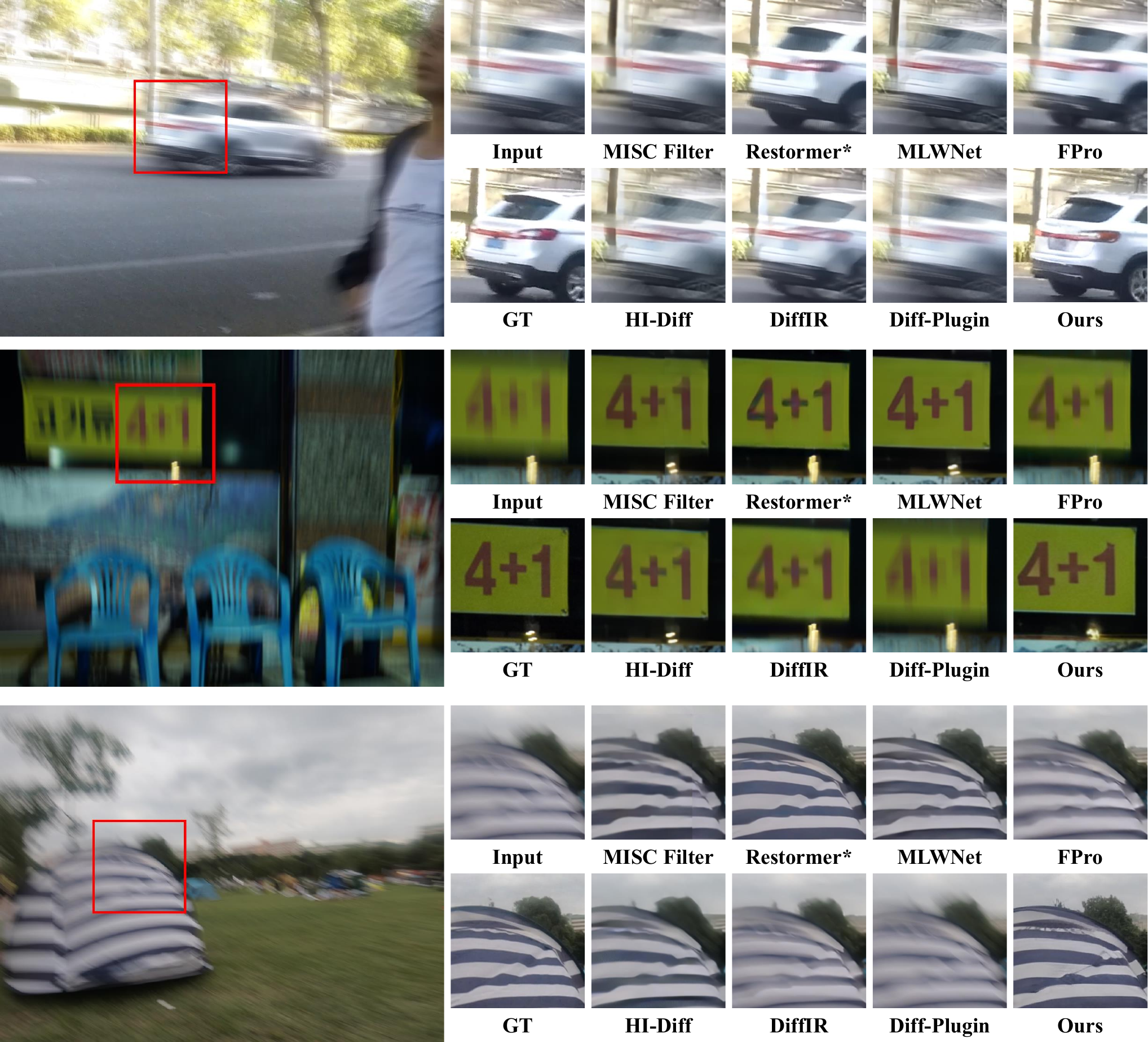}
    \captionof{figure}{Qualitative comparison on HIDE, Realblur, and REDS (From top to bottom). GLOWDeblur effectively handles diverse blur patterns with high-quality restorations.}
    \label{fig:comparison_supp_6datasets}
    
    \vspace{2em}
}]


\noindent 
To provide a more comprehensive assessment of our proposed method, we present additional visual results in this section, focusing on both cross-dataset generalization and real-world robustness.

First, to validate the generalization capability across different domains, Figure.~\ref{fig:comparison_supp_6datasets} illustrates qualitative comparisons on three distinct datasets: HIDE, Realblur, and REDS (arranged from top to bottom). As observed, GLOWDeblur effectively handles diverse blur patterns—ranging from camera shake to complex object motion—and restores fine-grained details with high fidelity, surpassing competing methods.

\begin{figure*}[!t]
    \centering
    \includegraphics[width=0.9\linewidth]{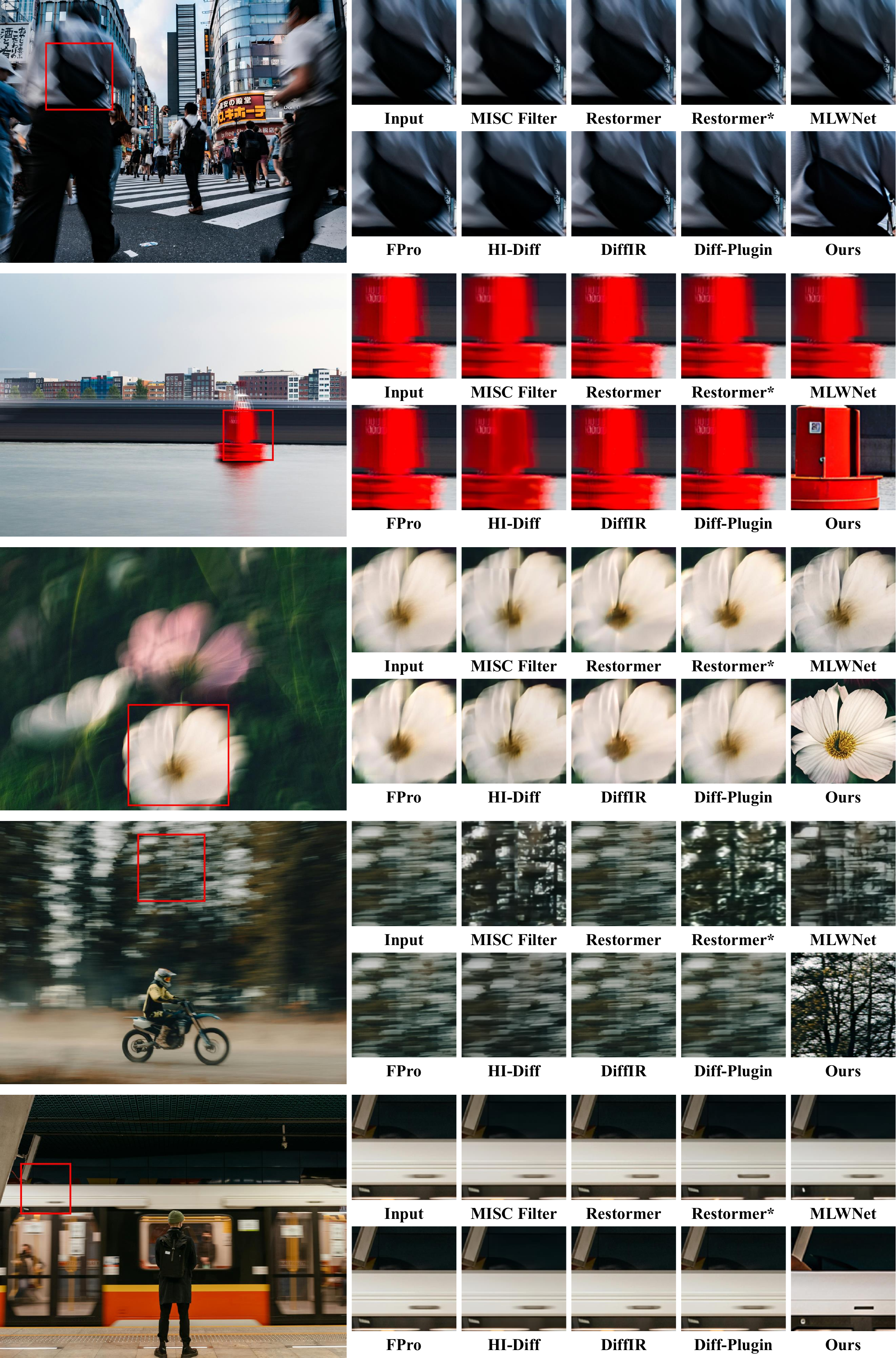}
    \caption{Comparison with SOTA deblur methods across real-world datasets. }
    \label{fig:comparison_supp_real_1}
\end{figure*}

\begin{figure*}[!t]
    \centering
    \includegraphics[width=0.9\linewidth]{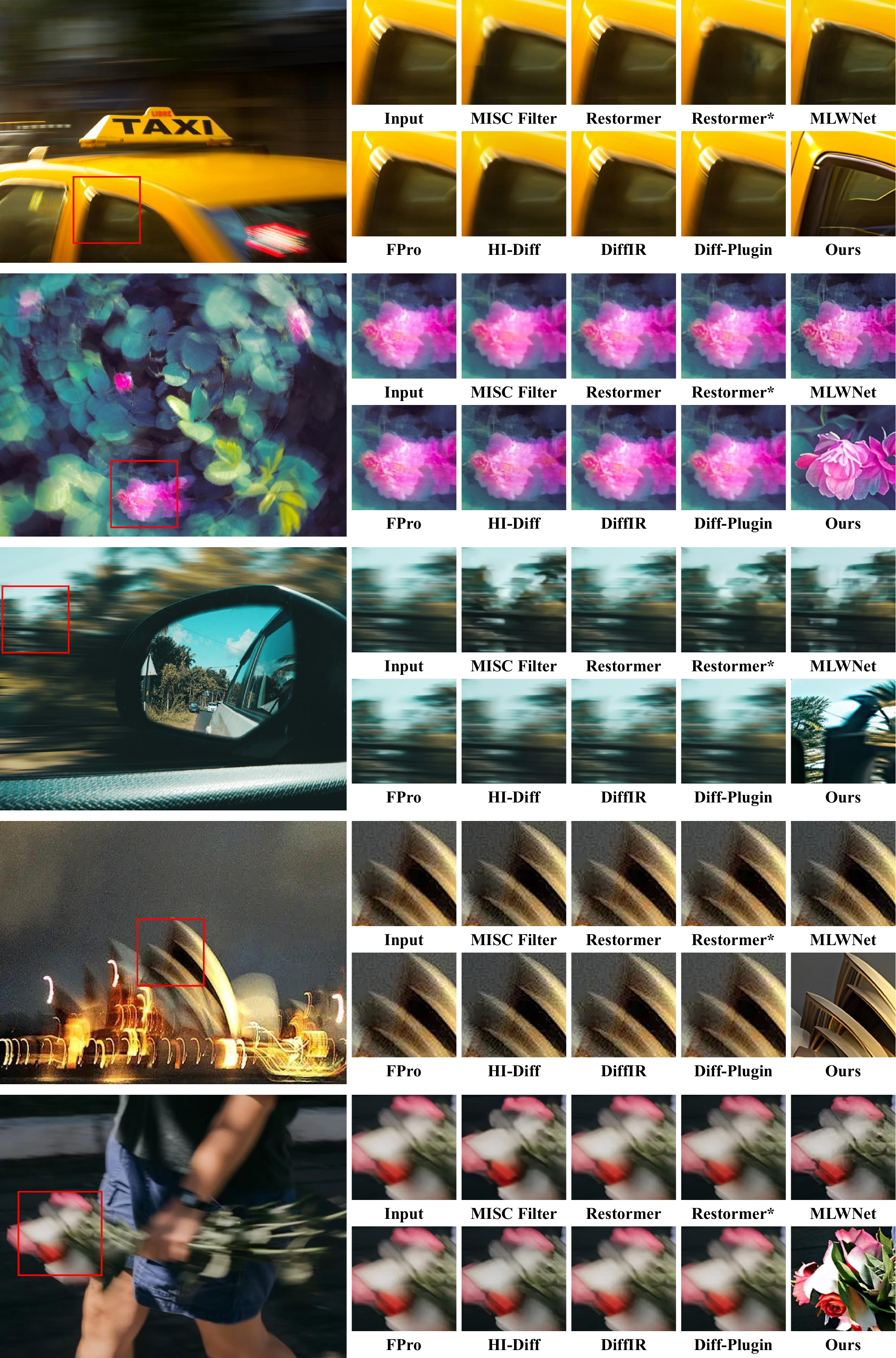}
    \caption{Comparison with SOTA deblur methods across real-world datasets. }
    \label{fig:comparison_supp_real_2}
\end{figure*}

Furthermore, we extend our evaluation to challenging real-world scenarios. Figure.~\ref{fig:comparison_supp_real_1} and Figure.~\ref{fig:comparison_supp_real_2} provide extensive comparisons with state-of-the-art methods. In these cases, where degradations are severe and unpredictable, our model demonstrates superior robustness, consistently recovering sharp structures and legible text while suppressing common restoration artifacts.



\clearpage
\twocolumn[{%

    \begin{multicols}{2}
    \subsection{RWBlur400}
    \label{sec:appendix_RWBlur400_details}
    \vspace{5pt}
        Existing motion deblurring datasets, such as GoPro~\cite{gopro}, HIDE~\cite{hide}, RealBlur~\cite{realblur}, REDS~\cite{reds}, and RSBlur~\cite{rsblur}, are predominantly captured in street environments. While valuable, they are inherently limited in their coverage of scene categories and blur patterns. Real-world blur, however, manifests in highly complex and diverse scenarios beyond street views. To address this gap and rigorously evaluate model generalization in the wild, we collected \textbf{RWBlur400}, a dataset comprising 400 diverse images sourced from the web. Unlike previous benchmarks, RWBlur400 encompasses a broad spectrum of subjects—ranging from views inside moving vehicles and dynamic wildlife to intricate flora—thereby posing a more realistic challenge. Representative examples are visualized in Figure.~\ref{fig:RWBlur400}.
    \end{multicols}

    \vspace{5pt} 

    \centering
    \includegraphics[width=1.0\linewidth, height=0.7\textheight, keepaspectratio]{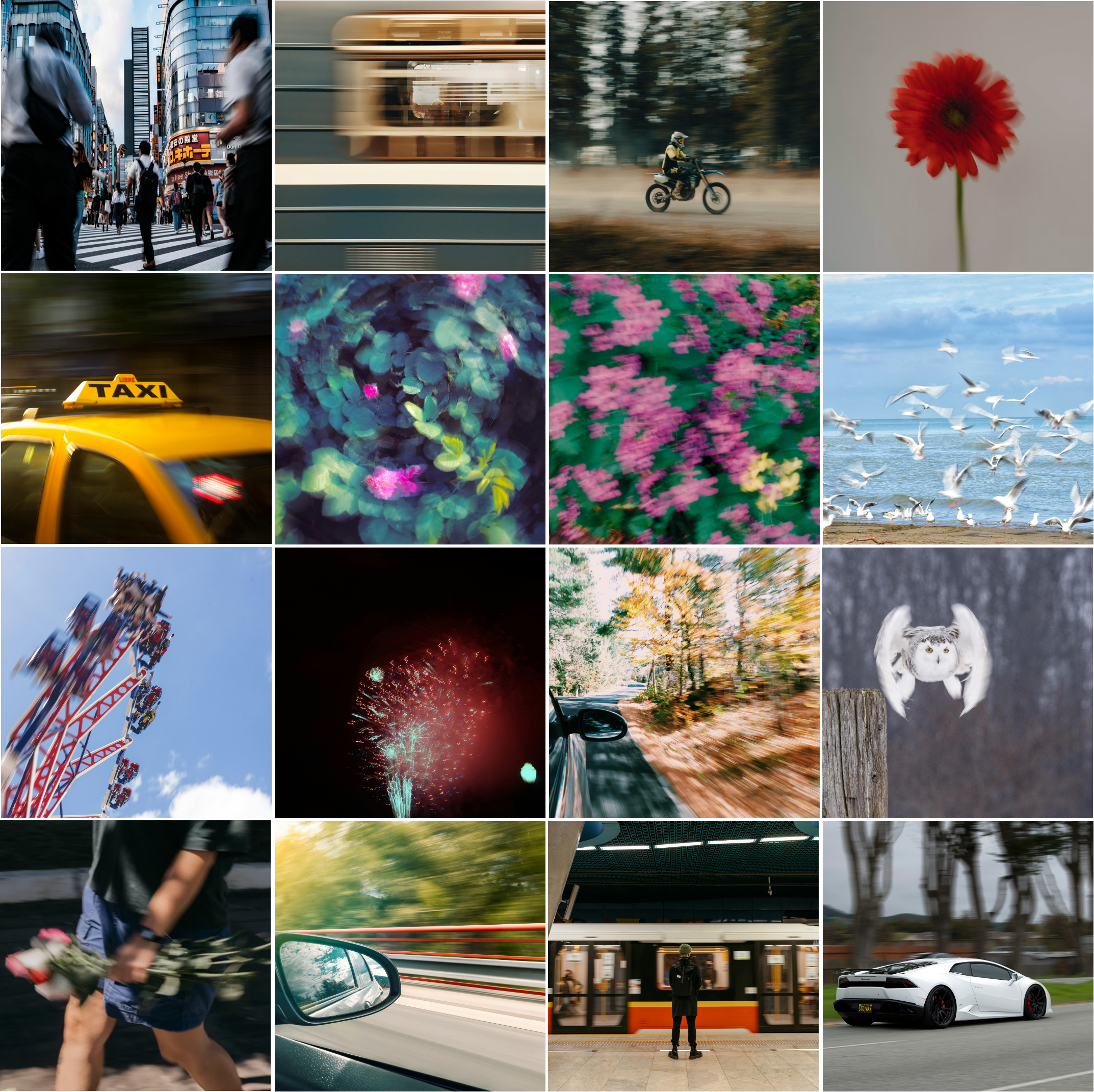}
    
    \captionof{figure}{\textbf{Visual samples from the collected RWBlur400 dataset.} Unlike existing benchmarks that are predominantly limited to street views, RWBlur400 covers a wide spectrum of semantic scenarios (e.g., natural landscapes, animals, and night scenes) and diverse blur patterns (e.g., rapid object motion, camera shake, and light trails), reflecting complex degradations in the wild.}
    \label{fig:RWBlur400}
    
}]

\end{document}